\documentclass{article}

\usepackage[preprint]{neurips_2025}

\usepackage[utf8]{inputenc} 
\usepackage[T1]{fontenc} 
\usepackage{hyperref} 
\usepackage{url} 
\usepackage{booktabs} 
\usepackage{amsfonts} 
\usepackage{nicefrac} 
\usepackage{microtype} 
\usepackage{xcolor} 
\usepackage{graphicx}
\usepackage{float}
\usepackage{amsmath}
\usepackage{cleveref}
\usepackage{subfig}
\usepackage{svg}
\usepackage{natbib}
\usepackage[switch]{lineno}
\usepackage{enumitem}
\usepackage{multirow}
\usepackage{wrapfig}
\usepackage{tabularx}

\title{Rethinking Intelligence: Brain-like Neuron Network}

\author{%
Weifeng Liu \\
Vista Zenith\\
\texttt{weifeng.liu@vistazenith.com} \\
}

\begin{document}
    \maketitle
    \vspace{-20pt}
    \begin{figure}[H]
        \centering
        \includegraphics[width=1.0\textwidth]{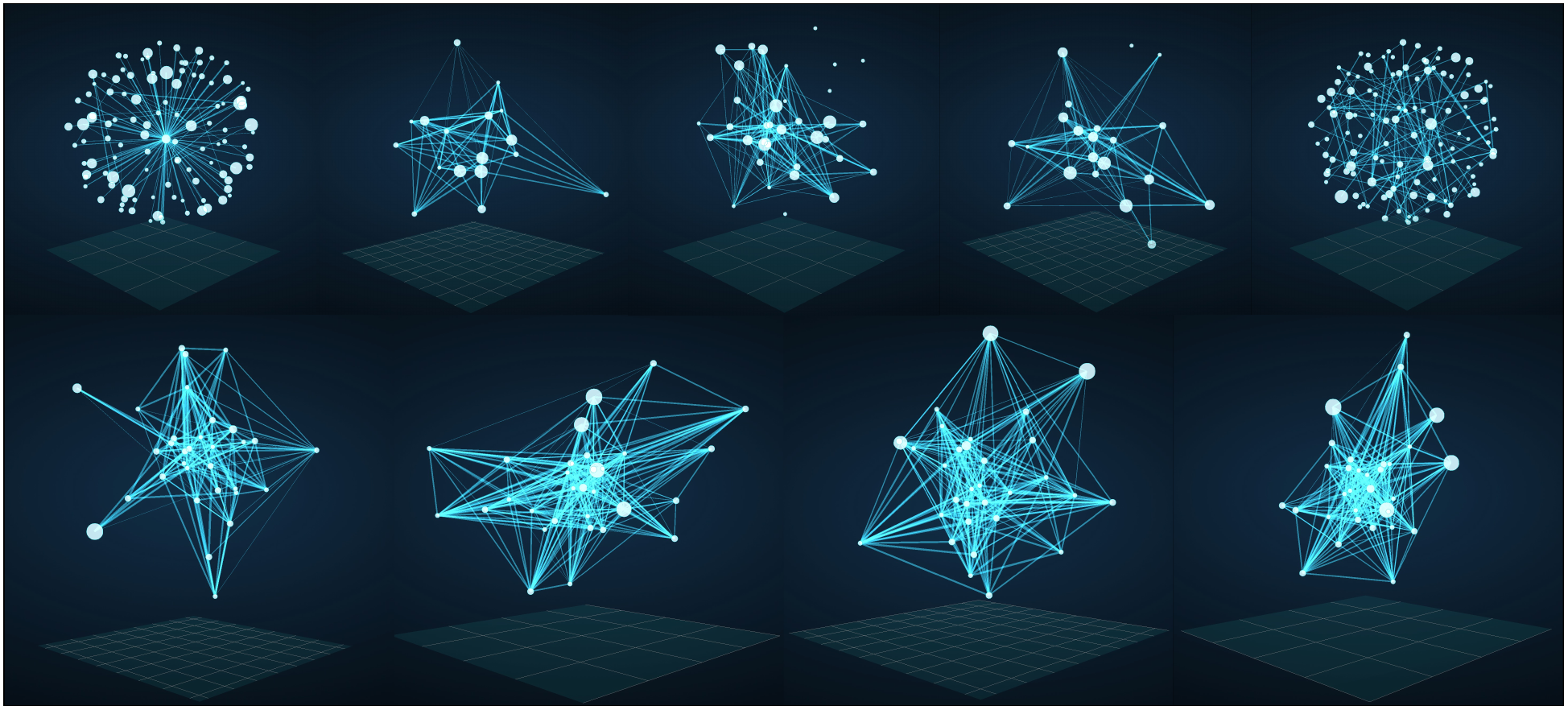}
        \caption{Visualization of the self-construced, self-evolved Brain-like Neuron
        Network architectures.}
        \label{fig:headline}
    \end{figure}

    \vspace{-10pt}
    \begin{abstract}
        \vspace{-5pt}
        Since their inception, artificial neural networks have relied on
        manually designed architectures and inductive biases to better adapt to
        data and tasks \cite{inductive-bias}. With the rise of deep learning and
        the expansion of parameter spaces, they have begun to exhibit brain-like
        functional behaviors \cite{ai-brain-like-functions}. Nevertheless,
        artificial neural networks remain fundamentally different from
        biological neural systems in structural organization, learning mechanisms,
        and evolutionary pathways \cite{brain_inspired_learning_review}.

        From the perspective of neuroscience, we rethink the formation and evolution
        of intelligence and proposes a new neural network paradigm, Brain-like
        Neural Network (BNN). We further present the first instantiation of a BNN
        termed LuminaNet that operates without convolutions or self-attention and
        is capable of autonomously modifying its architecture. We conduct
        extensive experiments demonstrating that LuminaNet can achieve self-evolution
        through dynamic architectural changes. On the CIFAR-10 \cite{cifar10},
        LuminaNet achieves top-1 accuracy improvements of 11.19\%, 5.46\% over LeNet-5
        \cite{lenet} and AlexNet \cite{alexnet}, respectively, outperforming MLP-Mixer
        \cite{mlp-mixer}, ResMLP \cite{resmlp}, and DeiT-Tiny \cite{deit} among MLP/ViT
        architectures. On the TinyStories \cite{TinyStories} text generation
        task, LuminaNet attains a perplexity of 8.4, comparable to a single-layer
        GPT-2-style Transformer \cite{gpt2,transformer}, while reducing computational
        cost by approximately 25\% and peak memory usage by nearly 50\%. Code
        and interactive structures are available at
        \url{https://github.com/aaroncomo/LuminaNet}.
    \end{abstract}

    \section{Introduction}
    \vspace{-3pt}
    \label{sec:intro}

    The brain is the most complex and efficient information processing system
    currently known to humankind, capable of realizing multi-level cognitive functions
    including perception, learning, reasoning, and creativity while under strict
    constraints of energy and structural resources. Since the introduction of artificial
    neural networks, researchers have long sought to build artificial
    intelligence systems inspired by the brain. In particular, with the rise of
    deep learning and large language models, artificial intelligence systems have
    achieved performance surpassing traditional algorithms in domains such as image
    recognition and natural language processing \cite{nlp}.

    However, mainstream artificial neural networks differ substantially from the
    real brain in their implementation paradigms. The biological brain is not a static,
    uniformly trained network, but rather a complex dynamical system that gradually
    evolves during development and remains continuously plastic across multiple time
    scales \cite{brain_inspired_learning_review}. In contrast, modern artificial
    neural networks typically rely on manually designed architectures,
    explicitly inductive biases, and centralized parameter update mechanisms.
    Their learning process amounts to searching for an optimal set of parameters
    within a fixed, bounded parameter space to fit the data distribution \cite{concept_synaptic_diversity},
    with structural constraints and operational assumptions that are
    fundamentally different from those of biological neural systems. From the perspectives
    of neuroscience and brain science, this paper reconsiders the origins and
    evolution of intelligence, proposing a novel artificial neural network paradigm,
    Brain-like Neural Network (BNN), that more closely mirrors the structure and
    evolutionary dynamics of biological brains. We also propose LuminaNet, a concrete
    implementation conforming to this paradigm.

    LuminaNet targets the retinotectal pathway to V1 visual cortex \cite{retina-to-V1,
    lgn, lgn-to-v1, magno} as its biological inspiration. It does not require
    predefined architectures, avoids convolution kernels and attention mechanisms,
    does not inject inductive biases, and instead defines only evolutionary rules.
    During training, neurons within the network autonomously modify their topological
    structure through four fundamental operations: splitting, growing,
    connecting, and pruning to achieve self-evolution. We observe that the
    network spontaneously emerges stable feedforward, feedback, and recurrent
    connections during growth, self-organizing into depth and width structures,
    and enhancing performance through the resulting complex topological
    architecture.

    Through self-evolution, LuminaNet achieves performance on CIFAR-10 \cite{cifar10}
    without any training tricks: it surpasses the classic convolutional
    architectures LeNet-5 \cite{lenet} and AlexNet \cite{alexnet} by 11.19\% and
    5.46\% in top-1 accuracy, respectively, and achieves a top-5 accuracy of
    98.09\%, outperforming MobileViT \cite{mobilevit} (97.39\%) and approaching ResNet-18
    \cite{resnet} (99.41\%) which has ten times more parameters. On the TinyStories
    text generation task, LuminaNet achieves a best PPL of 8.4 and top-1
    accuracy of 53.38\% without using self-attention, positional encoding, causal
    masking, or any training tricks, comparable to single-layer GPT-2-style Transformer
    \cite{gpt2,transformer} (PPL: 8.08, top-1: 53.29\%).

    Our contributions can be summarized as follows:
    \vspace{-3pt}
    \begin{itemize}[leftmargin=*, labelsep=1em]
        \item We propose a novel artificial neural network paradigm, \textbf{B}rain-like
            \textbf{N}eural \textbf{N}etworks (\textbf{BNN}), that more closely
            aligns with the structural and evolutionary mechanisms of biological
            brains.

        \item We present the first implementation of a self-architectural-evolving
            network, LuminaNet, that is convolution-free, attention-free, and bias-free.
            It can be applied to both image recognition and text generation
            tasks with minimal or no adjustments.

        \item LuminaNet employs explicit connection method, enabling high
            interpretability in terms of architectural evolution, internal data flow,
            and information propagation strength.

        \item We provide the first empirical demonstration that artificial neural
            networks can autonomously construct and optimize themselves without any
            human interventions. We also demonstrate that the key to self-evolution
            lies in the complex topological structures formed by neuron
            connections.
    \end{itemize}

    \begin{figure*}
        \centering
        \includegraphics[width=1.0\textwidth]{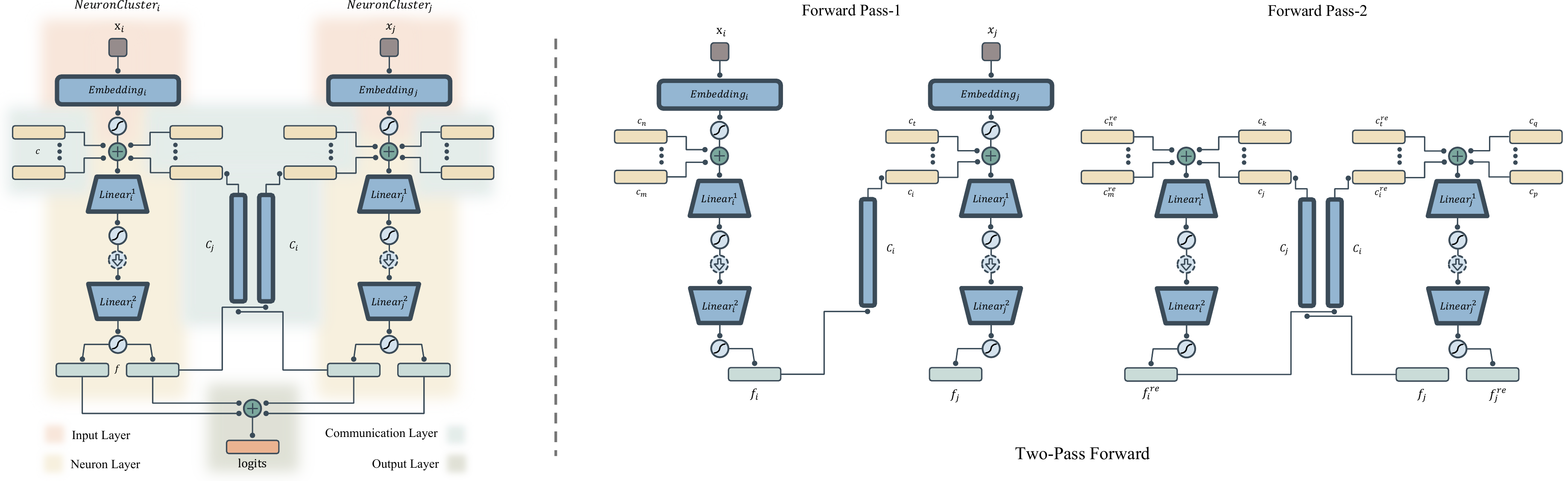}
        \caption{Structure of the Neuron Cluster (left) and Two-Pass Forward
        method (right).}
        \label{fig:pipeline}
    \end{figure*}

    \vspace{-3pt}
    \section{Defination}
    \vspace{-3pt}
    \label{sec:defination}

    BNN is not a specific architecture but rather a new paradigm that can be formally
    defined as follows:
    \vspace{-3pt}
    \begin{enumerate}[leftmargin=*, labelsep=1em]
        \item \textbf{Non–manually designed architectures:} Neuron structures
            that emerge through adaptive or self-organizing processes rather
            than explicit human design.

        \item \textbf{Structural assumptions similar to those of the brain:}
            Architectural limitations and computational principles that are
            aligned with biological neural systems.

        \item \textbf{Capability of autonomous evolution:} Including but not
            limited to architectural adjustments, self-organization of width and
            depth, spontaneous formation of connections, and the ability to maintain
            or enhance performance through evolutionary processes.

        \item \textbf{High interpretability}: The network exhibits high interpretability
            in terms of structural evolution, data flow, and information propagation.
    \end{enumerate}

    Formally, a Brain-like Neuron Network can be decomposed into four
    hierarchical layers:
    \vspace{-3pt}
    \begin{enumerate}[leftmargin=*, labelsep=1em]
        \item \textbf{Input Layer (IL)}: Embeds the original inputs into the latent
            space.

        \item \textbf{Neuron Layer (NL)}: Extracts features and do integration;
            neurons within this layer can grow.

        \item \textbf{Communication Layer (CL)}: Formally a single layer, but
            physically interwoven within the Neuron Layer. It contains feedforward,
            feedback, and recurrent connections, simulating synaptic
            communication between Neuron Clusters. It enables the exchange of excitatory
            or inhibitory signals \cite{vogels2011inhibitory} across different Neuron
            Clusters.

        \item \textbf{Output Layer (OL)}: Integrates outputs from all Neuron Clusters
            to produce the final result.
    \end{enumerate}

    \vspace{-3pt}
    \section{LuminaNet}
    \vspace{-3pt}
    \label{sec:luminanet}

    In biological neural systems, neurons with similar biophysical functions
    form Neuronal Populations \cite{neuron-cluster, microcolumn, minicolumn}, to
    collaboratively accomplish specific tasks. We emulate this phenomenon and
    design the only manually constructed foundational module in LuminaNet: the Neuron
    Cluster (NC).

    LuminaNet adheres to the four-layer formalized architecture of BNN.
    Communication between clusters occurs through the CL, which employs a fixed dimension
    $d_{hidden}$ as the channel. Each Neuron Cluster spans both the IL and the
    NL, receiving information from other clusters via the CL. Figure~\ref{fig:pipeline}
    illustrates the most complex possible connection pattern between two
    clusters.

    We divide RGB images channel-wise into $16 \times 16$ patches, and encode and
    integrate each patch using an independent Neuron Cluster.

    \subsection{Neuron Cluster}

    Given a Neuron Cluster $NC_{i}$ and input $x_{i}$, it performs the following
    computation in the IL to embed the raw input into a feature space and forward
    it to the NL. Here, $\sigma(\cdot)$ denotes the activation function:
    \begin{small}
        \begin{equation}
            e_{i}= \sigma\left( \text{Linear}_{i}\left( \text{Flatten}(x_{i}) \right
            ) \right)
            \vspace{-6pt}
        \end{equation}
    \end{small}

    Each part of the Neuron Cluster that is in the Neuron Layer supports three evolutionary
    mechanisms:

    \textbf{Splitting.} The original Neuron Cluster retains only the first half
    of its weights and biases, while the second half is separated to form a new one.

    \textbf{Growth.} The Neuron Cluster undergoes horizontal expansion or
    reduction, acquiring $n$ new neurons.

    \textbf{Connection.} Each Neuron Cluster $NC_{i}$ can establish multiple
    connections from other clusters. With a weight matrix $W_{C_j}$, each
    connection $C_{j}$ applies a linear transformation to the original
    activation signal $f_{j}$, mimicking the modulation of signal strength during
    synaptic transmission in biological brains. Suppose $NC_{i}$ has $n$
    incoming connections from other clusters, indexed in ascending order as the set
    $\mathcal{N}_{i}$. The aggregated signal received by $NC_{i}$ is then:
    \begin{small}
        \begin{equation}
            s_{i}= \frac{1}{n}\sum_{j \in \mathcal{N}_i}c_{j}= \frac{1}{n}\sum_{j
            \in \mathcal{N}_i}W_{C_j}f_{j}, \quad |\mathcal{N}_{i}| = n
            \vspace{-6pt}
        \end{equation}
    \end{small}

    \subsection{Two-Pass Forward}
    In biological brains, information flow is not limited to feedforward transmission
    but involves extensive feedback regulation and recurrent structures \cite{retina-to-V1,
    lgn, corticothalamic-feedback, hypo-feedback,feedback,recurrent-fund,
    recurrent}. As shown in Figure~\ref{fig:pipeline}, we propose a novel serial
    forward propagation mechanism called Two-Pass Forward, enabling LuminaNet to
    dynamically form feedforward, feedback, and recurrent connections through
    its evolving topology.

    Let $\mathcal{N}_{j}^{f}$ and $\mathcal{N}_{j}^{b}$ denote the ascendingly
    ordered sets of feedforward and feedback connection indices for cluster
    $NC_{j}$, respectively, satisfying:
    \begin{small}
        \begin{equation}
            \{\mathcal{N}_{j}^{f},\ \mathcal{N}_{j}^{b}\mid \max(\mathcal{N}_{j}^{f}
            )<j,\ \min (\mathcal{N}_{j}^{b})>j\}
        \end{equation}
    \end{small}

    \textbf{Pass-1.} Traverse all clusters. For each cluster $NC_{j}$, compute its
    Pass-1 output using:
    \begin{small}
        \begin{equation}
            f_{j}=\text{NL}_{j}\left(\frac{e_{j}+\sum_{i\in\mathcal{N}_j^f}W_{C_i}f_{i}}{1+|\mathcal{N}_{j}^{f}|}
            \right)
        \end{equation}
    \end{small}

    \textbf{Pass-2.} This is a dynamic forward process. Traverse all clusters.
    At the time of visiting $NC_{j}$, define the set of valid feedforward connections
    as follows:
    \begin{small}
        \begin{equation}
            \tilde{\mathcal{N}}_{j}^{f}= \{\, i \mid\ i \in \mathcal{N}_{j}^{f},
            \max (\mathcal{N}_{j}^{f})<j,\ f_{i}^{re}\text{ exists}\,\}
        \end{equation}
    \end{small}

    Then, compute the Pass-2 output for $NC_{j}$ as:
    \begin{small}
        \begin{equation}
            f_{j}^{re}=\text{NL}_{j}\left(\frac{\sum_{i\in\tilde{\mathcal{N}}_j^f}W_{C_i}f_{i}^{re}+
            \sum_{k\in\mathcal{N}_j^b}W_{C_k}f_{k}}{|\tilde{\mathcal{N}}_{j}^{f}|+|\mathcal{N}_{j}^{b}|}
            \right)
        \end{equation}
    \end{small}

    \textbf{Integration.} Let $\mathcal{O}_{j}$ denote the final valid output set
    for cluster $NC_{j}$ after Two-Pass Forward:
    \begin{small}
        \begin{equation}
            \mathcal{O}_{j}= \left\{ f_{j}\right\} \;\cup\; \left\{ f_{j}^{\mathrm{re}}
            \;\middle|\; f_{j}^{\mathrm{re}}\ \text{exists}\right\}
        \end{equation}
    \end{small}

    The network integrates the outputs $\mathcal{O}_{j}$ from all $n$ clusters,
    and produces the final prediction $\bar o$:
    \begin{small}
        \begin{equation}
            \bar o = \frac{ \sum_{j=1}^{n}\sum_{o \in \mathcal{O}_j}o }{ \sum_{j=1}^{n}|\mathcal{O}_{j}|
            }
        \end{equation}
    \end{small}

    \subsection{Strategies}
    \label{sec:strategies} LuminaNet defines four fundamental evolutionary
    strategies, each selected with independent relative probabilities $\{p_{s}, p
    _{g}, p_{c}, p_{p}\}$ during evolution. During training, we collect $n$ features
    from the NL of each NC, and use their variance $v$ as the basis for strategy
    selection. A higher variance $v$ for a Neuron Cluster indicates stronger information
    separation capability and greater contribution to the final output \cite{jolliffe2002principal,
    bengio2013representation,fisher1936use}.

    \textbf{Split Strategy.} We select a cluster whose variance exceeds the
    $\alpha$-quantile, and split it to mitigate overfitting risk and promote more
    balanced capacity distribution across clusters. We set $\alpha$ to 0.9.

    \textbf{Grow Strategy.} We select a cluster whose variance is below the $\beta$-quantile,
    and increase the number of neurons in its Neuron Layer to enhance its
    capacity. We set $\beta$ to 0.4.

    \textbf{Connect Strategy.} High-variance clusters transmit information to
    low-variance clusters via connections. For two clusters indexed $i < j$, if
    $v_{i}> v_{j}$, a feedforward connection $i \to j$ is established; otherwise,
    a feedback connection $j \to i$ is formed.

    \textbf{Prune Strategy.} For a connection $C$, a larger weight norm $\|W_{C}\|
    _{F}$ indicates higher sensitivity to input and a broader range of linear transformations,
    enabling stronger transmission
    \cite{bartlett2017spectrally, golub1996matrix,poole2016exponential} of
    excitatory or inhibitory signals. We use the Frobenius norm \cite{golub1996matrix}
    as a proxy for connection strength \cite{neyshabur2015path} and compute threshold
    as follows:
    \begin{small}
        \begin{equation}
            th_{j}\;=\; \theta\cdot \frac{\sum_{i\in\mathcal{N}_j^f\cup\mathcal{N}_j^b}\left\|
            W_{C_i}\right\|_{F}}{|\mathcal{N}_{j}^{f}\cup\mathcal{N}_{j}^{b}|}
        \end{equation}
    \end{small}

    where $\theta$ controls the pruning intensity; we set it to 0.9. Connections
    within $NC_{j}$ with weight norms below $th_{j}$ are pruned.

    \subsection{Optimization and Evolution}
    Both optimization and evolution aim to minimize the cross-entropy loss. When
    the loss fails to decrease for more than a predefined patience threshold, the
    network triggers an evolutionary event, modifying its architecture
    autonomously to explore new directions.

    \vspace{-3pt}
    \section{Image Recognition}
    \vspace{-3pt}

    \subsection{Setup}
    \vspace{-2pt}
    \label{sec:setup-cifar10}

    We use CIFAR-10 for image recognition and evaluate three categories of
    models with comparable parameter counts: 1) CNNs: LeNet-5 \cite{lenet},
    AlexNet \cite{alexnet}, MobileNet-V2/V3 \cite{mobilenetv2, mobilenetv3},
    DenseNet \cite{densenet}, PreAct-ResNet \cite{preresnet}, ResNet
    \cite{resnet}; 2) ViTs: MobileViT \cite{mobilevit}, DeiT-Tiny \cite{deit}; 3)
    MLPs: ResMLP \cite{resmlp}, gMLP \cite{gmlp}, MLP-Mixer \cite{mlp-mixer}.
    These models have covered mainstream network paradigms.

    LuminaNet is trained using AdamW \cite{adamw} with a learning rate of 0.001,
    weight decay of 0.05, and a batch size of 1024. To isolate the intrinsic
    capacity of each architecture from the advantages conferred by training
    tricks, we remove all learning-rate schedulers used in the baselines, while
    keeping other hyperparameters consistent with the original papers. For MobileNet
    V2 and V3, the learning rate is adjusted to 0.001 due to training
    instability.

    Data augmentation is applied, and all baselines are trained until the training
    accuracy exceeds 0.99, with the best performance reported. Since LuminaNet continuously
    modifies its architecture for self-regularization and cannot reliably overfit
    to 0.99, we cap its training at 10,000 epochs.

    Training is conducted on Apple M2 and NVIDIA A800, while inference is evaluated
    on Snapdragon 8 Elite (mobile phone), Apple M2 (CPU), and NVIDIA A800 (GPU).

    \begin{table}[]
        \scriptsize
        \centering
        \caption{Baseline comparison for the image recognition task, sorted by Top-1
        accuracy rate.}
        \vspace{5pt}
        \label{tab:baseline-cifar10}
        \begin{tabular}{@{}lcccccc|ccc@{}}
            \toprule Model                     & Params (M)      & Structure & Evolve (Params)         & Top-1          & Top-3          & Top-5          & Rank-1      & Rank-3      & Rank-5     \\
            \midrule gMLP-15 \cite{gmlp}       & 1.60            & MLP       & -                       & 57.38          & 83.72          & 92.71          & 18          & 18          & 18         \\
            LeNet-5 \cite{lenet}               & 0.06            & CNN       & -                       & 62.81          & 86.87          & 94.69          & 17          & 17          & 17         \\
            MobileNet-V3 \cite{mobilenetv3}*   & 1.67            & CNN       & -                       & 63.16          & 89.63          & 96.72          & 16          & 15          & 11         \\
            MLP-Mixer \cite{mlp-mixer}         & 3.80            & MLP       & -                       & 64.99          & 89.92          & 96.56          & 15          & 14          & 12         \\
            DeiT-Tiny \cite{deit}              & 5.68            & ViT       & -                       & 66.81          & 89.20          & 95.93          & 14          & 16          & 14         \\
            AlexNet \cite{alexnet}             & 2.47            & CNN       & -                       & 68.54          & 90.24          & 96.28          & 13          & 13          & 13         \\
            \textbf{LuminaNet-10}              & \textbf{0.36}   & BNN       & 0.03 $\rightarrow$ 0.36 & \textbf{69.28} & \textbf{91.87} & \textbf{97.58} & \textbf{12} & \textbf{11} & \textbf{8} \\
            ResMLP-6 \cite{resmlp}             & 3.36            & MLP       & -                       & 71.53          & 92.19          & 97.28          & 11          & 10          & 10         \\
            \textbf{LuminaNet-128}             & \textbf{3.16}   & BNN       & 0.79 $\rightarrow$ 3.16 & \textbf{72.84} & \textbf{92.59} & \textbf{97.84} & \textbf{10} & \textbf{9}  & \textbf{6} \\
            \textbf{LuminaNet-32}              & \textbf{{}1.43} & BNN       & 0.12 $\rightarrow$ 1.43 & \textbf{73.12} & \textbf{92.98} & \textbf{97.83} & \textbf{9}  & \textbf{6}  & \textbf{7} \\
            \textbf{LuminaNet-64}              & \textbf{{}2.14} & BNN       & 0.30 $\rightarrow$ 2.14 & \textbf{73.58} & \textbf{93.29} & \textbf{97.86} & \textbf{8}  & \textbf{5}  & \textbf{5} \\
            \textbf{LuminaNet-84}              & \textbf{1.86}   & BNN       & 0.43 $\rightarrow$ 1.86 & \textbf{74.00} & \textbf{93.74} & \textbf{98.09} & \textbf{7}  & \textbf{4}  & \textbf{4} \\
            MobileViT \cite{mobilevit}         & \textbf{{}5.33} & ViT       & -                       & 75.62          & 92.95          & 97.39          & 6           & 7           & 9          \\
            PreAct-ResNet-20 \cite{preresnet}  & 0.27            & CNN       & -                       & 76.27          & 91.22          & 95.40          & 5           & 12          & 15         \\
            MobileNet-V2 \cite{mobilenetv2}*   & 2.24            & CNN       & -                       & 79.27          & 95.10          & 98.17          & 4           & 3           & 3          \\
            PreAct-ResNet-110 \cite{preresnet} & 1.73            & CNN       & -                       & 82.53          & 92.82          & 95.23          & 3           & 8           & 16         \\
            ResNet18 \cite{resnet}             & 11.69           & CNN       & -                       & 87.96          & 97.59          & 99.41          & 2           & 2           & 2          \\
            DenseNet-100BC \cite{densenet}     & 0.77            & CNN       & -                       & 90.06          & 98.32          & 99.61          & 1           & 1           & 1          \\
            \bottomrule
        \end{tabular}
    \end{table}

    \subsection{Baseline Comparison}
    As shown in Table~\ref{tab:baseline-cifar10}, LuminaNet demonstrates competitive
    performance across multiple parameter scales through self-evolution. Notably,
    among small networks, LuminaNet-10, using only 0.36M parameters, achieves a Top-1
    accuracy of 69.28\%, surpassing classical convolutional models such as LeNet-5
    and AlexNet. In the medium-parameter regime, LuminaNet variants with different
    $d_{\text{hidden}}$ outperforming MLP-based and ViT architectures such as MLP-Mixer,
    ResMLP-6, and DeiT-Tiny, while requiring significantly fewer parameters than
    MobileViT and ResNet-based models. When considering Top-5, \textbf{all} LuminaNets
    rank below only MobileNet-V2, ResNet18, and DenseNet-100BC.

    Most baselines exhibit substantial performance degradation after removing schedulers,
    with MobileNet-V2/V3 even experiencing training collapse. In contrast,
    LuminaNet maintains or improves performance even as it continuously modifies
    its architecture, highlighting its inherent architectural advantages. This
    experiment provides strong evidence that the self-evolving BNN paradigm, embodied
    by LuminaNet, possesses performance potential comparable to static deep
    learning architectures.

    \subsection{Ablation}

    \begin{table}[]
        \scriptsize
        \centering
        \caption{Research on the ablation of Neuron Clusters and their
        connections.}
        \vspace{5pt}
        \label{tab:cifar10-ablation}
        \setlength{\tabcolsep}{3pt}
        \begin{tabular}{@{}cccccccccccccc@{}}
            \toprule                & \multicolumn{4}{c|}{Raw} & \multicolumn{3}{c|}{A: new NCs \& all conns} & \multicolumn{3}{c|}{B: new NCs \& their conns} & \multicolumn{3}{c}{C: all conns} \\
            \cmidrule(l){2-14} Dim. & Clusters                 & Connections                                  & Params                                         & \multicolumn{1}{c|}{Top-1}      & Params & Top-1 & \multicolumn{1}{c|}{Perf. Gap} & Params & Top-1 & \multicolumn{1}{c|}{Perf. Gap} & Params & Top-1 & \multicolumn{1}{c}{Perf. Gap} \\
            \midrule 10             & 106                      & 119                                          & 0.36M                                          & 69.28                           & 0.04M  & 32.28 & -53.41                         & 0.05M  & 35.4  & -48.90                         & 0.75M  & 32.71 & -52.79                        \\
            32                      & 82                       & 120                                          & 1.43M                                          & 73.12                           & 0.21M  & 35.94 & -50.85                         & 1.31M  & 44.76 & -38.79                         & 0.23M  & 28.73 & -60.71                        \\
            64                      & 43                       & 103                                          & 2.14M                                          & 73.58                           & 0.66M  & 30.63 & -58.37                         & 0.79M  & 52.94 & -28.05                         & 1.72M  & 27.33 & -62.86                        \\
            84                      & 21                       & 119                                          & 1.86M                                          & 74.00                           & 0.63M  & 33.75 & -54.39                         & 1.07M  & 66.7  & -9.86                          & 1.02M  & 23.32 & -68.49                        \\
            128                     & 17                       & 82                                           & 3.16M                                          & 72.84                           & 1.45M  & 34.70 & -52.36                         & 2.24M  & 69.43 & -4.68                          & 1.81M  & 29.08 & -60.08                        \\
            \bottomrule
        \end{tabular}
    \end{table}

    Table~\ref{tab:cifar10-ablation} reports the impact of three different
    ablation settings on network performance.

    Case A: All connections are removed, and only the initial clusters are
    retained. Performance varies little across different $d_{\text{hidden}}$,
    indicating that individual clusters possess similar and limited capacity, and
    the network is unable to effectively aggregate local features extracted by different
    clusters.

    Case B: Only the initial clusters and their inter-cluster connections are
    retained. Since the initial clusters already cover all spatial locations in the
    input, the aggregated information through cluster-level connections leads to
    a significant performance boost compared to Case A. The improvement increases
    substantially as $d_{\text{hidden}}$ grows.

    Case C: All clusters are retained, but all connections are removed.
    Performance drops sharply, even often below that of Case A, for the network
    now contains more local features that cannot be integrated.

    Comparing these ablation experiments with the Case Raw reveals that the
    complex topological structure formed by connections is critical for
    performance enhancement, which is consistent with the physiological organization
    of the human brain.

    \subsection{Network Evolving}
    \begin{wrapfigure}
        {r}{0.45\textwidth}
        \vspace{-50pt}
        \centering
        \includegraphics[width=0.43\textwidth]{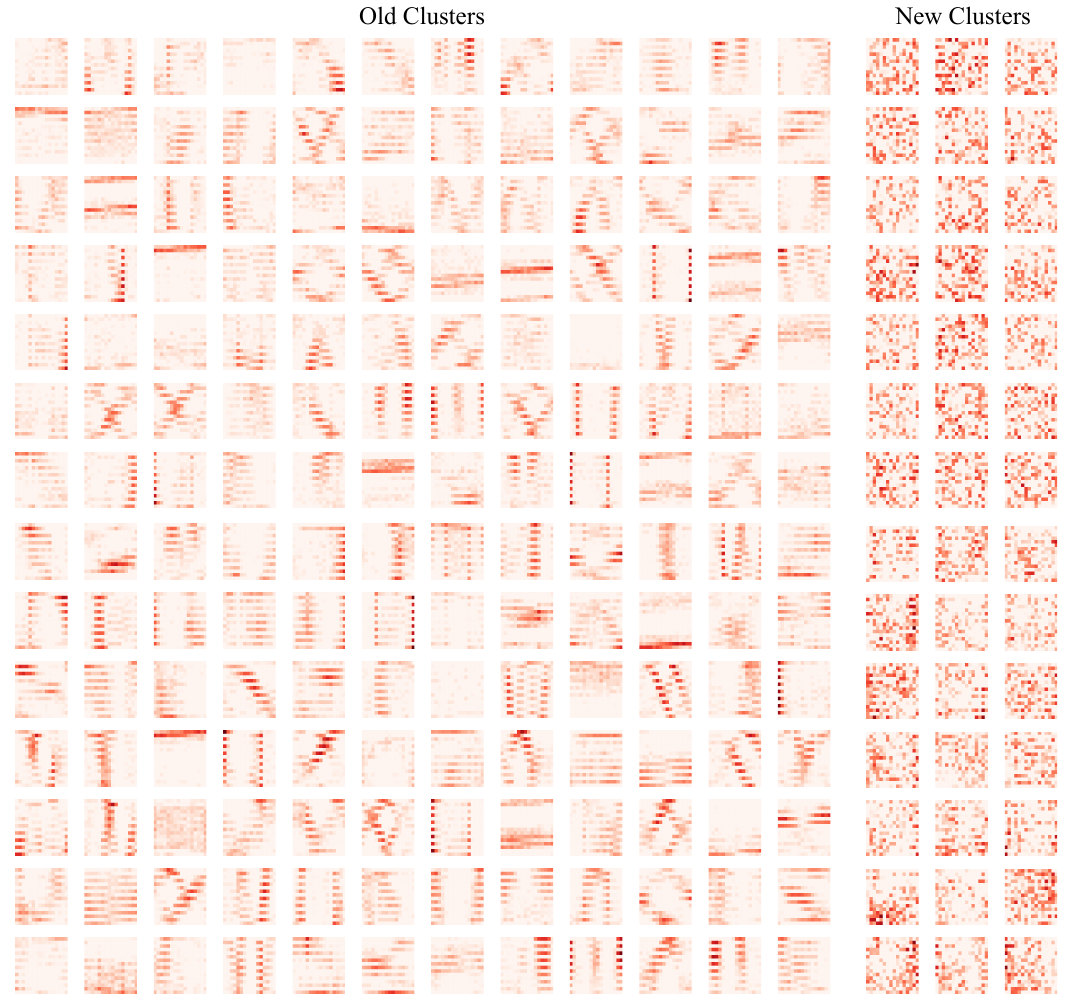}
        \caption{Visualization of the neuron weight matrices in the Input Layer.}
        \label{fig:embedding_weight}
        \vspace{-25pt}
    \end{wrapfigure}

    \begin{table}[]
        \centering
        \scriptsize
        \setlength{\tabcolsep}{3pt}
        \caption{The detailed evolution trajectories of several LuminaNets
        mentioned in Table~\ref{tab:baseline-cifar10}}
        \vspace{5pt}
        \label{tab:network-evolving}
        \begin{tabular}{@{}ccc@{\ $\rightarrow$\ }lc@{\ $\rightarrow$\ }lc@{\ $\rightarrow$\ }lcc
        r@{\ $\rightarrow$\ }l@{}}
            \toprule Dim. & Params (M)              & \multicolumn{2}{c}{Clusters} & \multicolumn{2}{c}{Connections} & \multicolumn{2}{c}{Topological Depth} & Max In-Degree & Cycles          & \multicolumn{2}{c}{Neurons (Min./Max./Avg.)} \\
            \midrule 10   & 0.03 $\rightarrow$ 0.36 & \ 12                         & 106                             & \ \ \ 0                               & 119           & \quad\quad\ \ 0 & 12                                          & 0 $\rightarrow$ 18 & 0 $\rightarrow$ 4 & 120 (5/20/10)     & 3640 (5/53/34)    \\
            32            & 0.12 $\rightarrow$ 1.43 & \ 12                         & 82                              & \ \ \ 0                               & 120           & \quad\quad\ \ 0 & 8                                           & 0 $\rightarrow$ 27 & 0 $\rightarrow$ 3 & 384 (16/64/32)    & 9728 (18/182/119) \\
            64            & 0.30 $\rightarrow$ 2.14 & \ 12                         & 43                              & \ \ \ 0                               & 103           & \quad\quad\ \ 0 & 8                                           & 0 $\rightarrow$ 25 & 0 $\rightarrow$ 8 & 768 (32/128/64)   & 7808 (32/352/181) \\
            84            & 0.43 $\rightarrow$ 1.86 & \ 12                         & 21                              & \ \ \ 0                               & 119           & \quad\quad\ \ 0 & 11                                          & 0 $\rightarrow$ 17 & 0 $\rightarrow$ 6 & 1008 (42/168/84)  & 3360 (64/256/160) \\
            128           & 0.79 $\rightarrow$ 3.16 & \ 12                         & 17                              & \ \ \ 0                               & 82            & \quad\quad\ \ 0 & 8                                           & 0 $\rightarrow$ 15 & 0 $\rightarrow$ 8 & 1536 (64/256/128) & 4864 (80/640/286) \\
            \bottomrule
        \end{tabular}
    \end{table}

    Table~\ref{tab:network-evolving} reports the evolutionary trajectories of five
    different variants across the baseline experiments. All networks
    autonomously evolved into architectures without any human design or
    intervention, featuring both increased width and depth, forming extensive
    feedforward, feedback, and recurrent connections, resulting in highly complex
    topological structures. In the case of $d_{\text{hidden}}= 32$, the total number
    of neurons even expanded to 25.3 times the initial count.

    \subsection{Input Layer Visualization}

    LuminaNet is a dynamic network, meaning that its Neuron Clusters do not undergo
    uniform optimization, this fundamentally differs from the standard training
    paradigm of current mainstream neural networks. As a result, LuminaNet
    simultaneously contains newly formed clusters and older clusters that have
    existed for extended periods. To investigate the preferred raw pixel features
    encoded by each cluster, we project the encoder weight matrices of the Input
    Layer back into the spatial patch shape and visualize them. As shown in Figure~\ref{fig:embedding_weight},
    after training, the cluster encoders exhibit distinct directional and shape preferences.
    In contrast, encoders of clusters that evolved later have not yet developed such
    specific preferences due to shorter training duration. However, their Neuron
    Layers can receive features extracted from lower-level clusters via connections
    and also receive feedback from higher-level clusters, thereby preventing
    network collapse that might otherwise arise from the random initialization
    of encoder parameters during evolution. We also provide visualizations of the
    evolutionary process in Appendix~\ref{supp:encoder-evolution}.

    \subsection{Neuron Layer Visualization}
    \begin{figure*}
        \centering
        \includegraphics[width=1.0\textwidth]{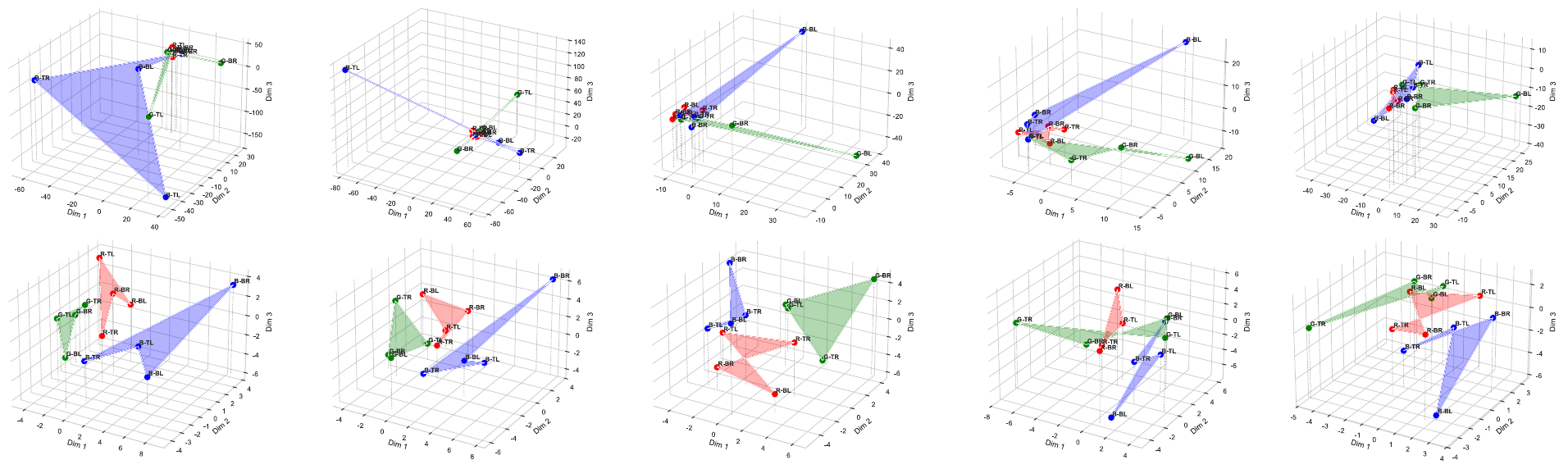}
        \caption{MDS visualizations of the features extracted by Neuron Layer
        after removing the connections (top) and before removing the connections
        (bottom). R, G, and B correspond to the red, green, and blue channels,
        respectively; T/B indicate the top and bottom, and L/R indicate the left
        and right.}
        \label{fig:mds}
    \end{figure*}

    The core of LuminaNet lies in its growable Neuron Layer and those dynamic
    connections injected into it. As shown in Figure~\ref{fig:mds}, when
    connections are removed, no information exchange remains within the network,
    causing the feature distribution in high dimension to become extremely
    skewed. In contrast, the presence of connections enables each cluster to
    access relative information from other patches, thereby preserving a structured
    geometric form in the representation space. This allows the network to
    efficiently encode the entire image within a more compact space. Additionally,
    it prevents different patches from collapsing into a single cluster in high-dimensional
    space, effectively maintaining semantic separability.

    Although MDS \cite{torgerson1952multidimensional,kruskal1964multidimensional}
    cannot directly prove linear separability across channels in latent space,
    the stable low-level manifold structure strongly suggests that channel and
    positional information is encoded in the Neuron Layer’s high-dimensional
    space in a low-complexity manner, i.e., the network can recover the original
    image via a specific topological structure in the high-dimensional space.
    This is analogous to the retinotopic mapping from the retina to the V1
    cortex \cite{retina-to-V1, lgn-to-v1, lgn}.

    \subsection{Connection Analyze}
    \begin{figure}
        \centering
        \includegraphics[width=1.0\textwidth]{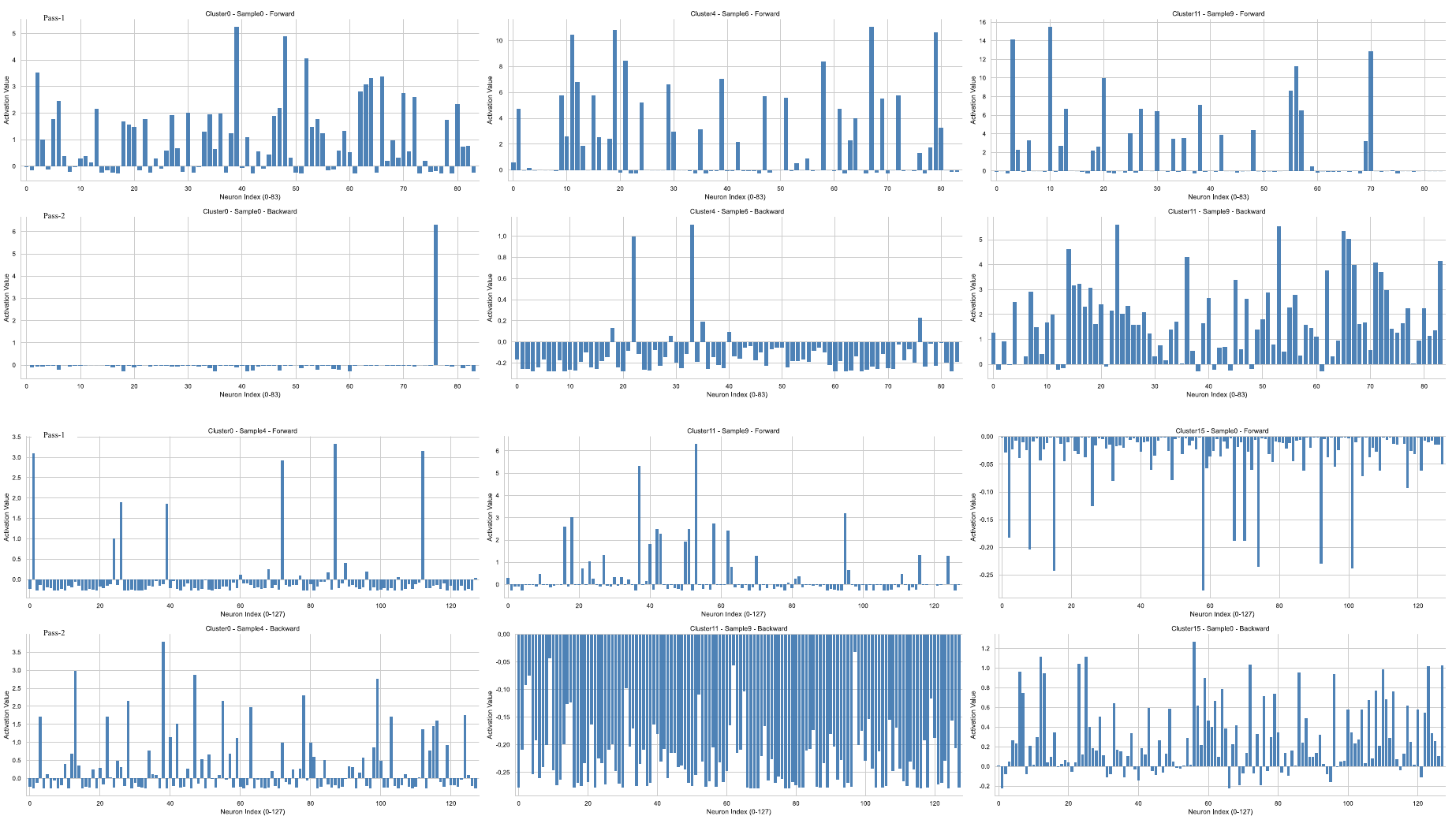}
        \caption{ Neuron-layer output activations. The x-axis denotes the neuron
        index and the y-axis denotes the activation magnitude. Forward indicates
        activations after Pass-1 (forward information only), whereas Backward
        indicates activations after Pass-2 (with feedback information).}
        \label{fig:activation}
    \end{figure}

    To further investigate the effect of the connections on the network, we
    visualize the output activations of the neuron layer. As shown in Figure~\ref{fig:activation},
    the activation values before and after feedback indicates most neurons
    exhibit noticeable changes in activation magnitude after feedback; some neurons
    become inactive, and some even undergoes polarity reversal in their
    activations. This provides strong evidence that the connections autonomously
    formed by the network can transmit excitatory and inhibitory signals of varying
    intensities effectively.

    \subsection{Structure Visualization}
    \label{sec:cifar10-structure}
    \begin{figure*}
        \centering
        \includegraphics[width=1.0\textwidth]{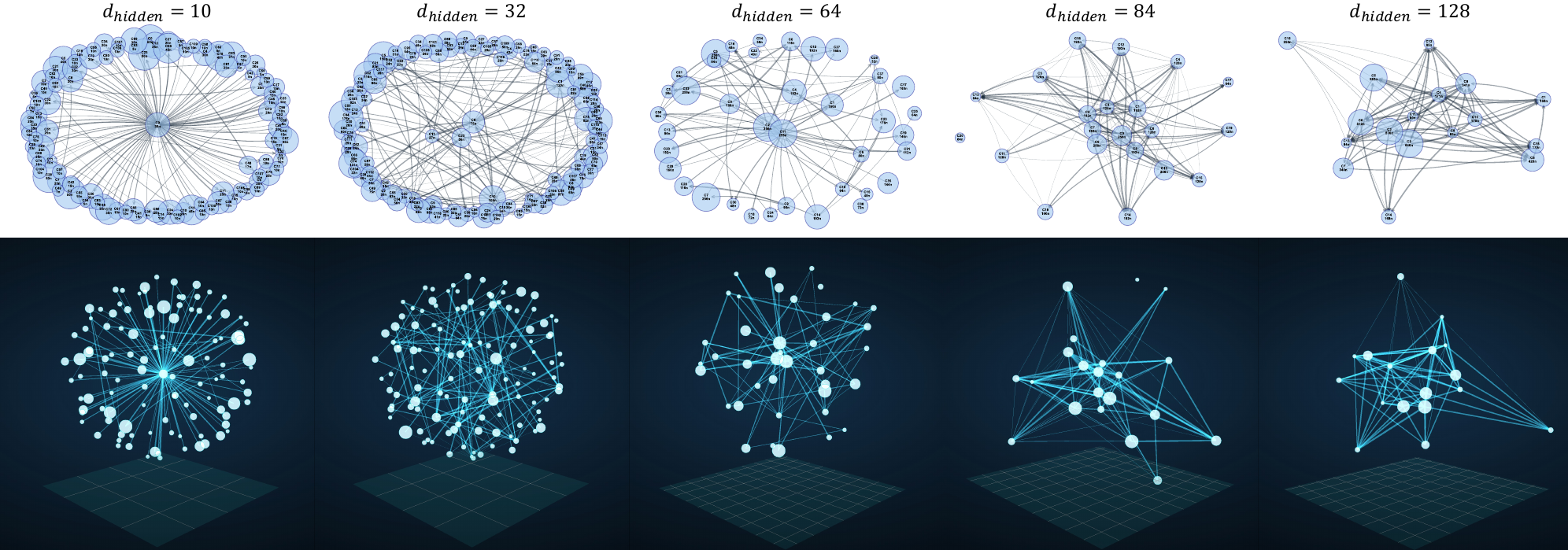}
        \caption{Architectures autonomously formed by LuminaNet on the CIFAR-10
        task. The thickness of the connecting lines indicates the strength of the
        connections.}
        \label{fig:structure-cifar10}
    \end{figure*}

    The evolution of LuminaNet is not a stable convergence process but rather
    leads to topological architectures of diverse morphologies. Crucially, the
    final structure and its degree of evolution are highly sensitive to the
    strategy probabilities $\{p_{s}, p_{g}, p_{c}, p_{p}\}$. Figure~\ref{fig:structure-cifar10}
    presents the architecture of several LuminaNets trained on CIFAR-10 task,
    visualized in both 2D and 3D spaces.

    For networks with small $d_{\text{hidden}}$, the hidden layer dimension becomes
    a bottleneck, prompting the network to rely on splitting to generate more
    clusters, thereby enabling feature extraction across diverse latent spaces and
    forming extensive connections to aggregate information in width. In contrast,
    networks with larger $d_{\text{hidden}}$ tend to evolve deeper hierarchical structures
    and recurrent connections, enhancing the network’s reusability and
    representational capacity. We will present quantitative analysis regarding connections
    in Appendix~\ref{supp:connection-weights}.

    \vspace{-3pt}
    \section{Text Generation}
    \vspace{-3pt}
    \label{supp:text-generation}

    Although the network was not originally proposed for autoregressive tasks, its
    architecture featuring feedforward, feedback, and recurrent connections, can
    inherently handle such problems \cite{mienye2024recurrent,wen2018deep,rnn}. We
    conducted experiments on the TinyStories task not to pursue optimal generation
    performance, but to explore whether a self-evolving network without self-attention
    \cite{transformer}, positional encoding \cite{transformer}, causal masking
    \cite{transformer} and state memory \cite{rnn, lstm} can effectively learn
    temporal dependencies.

    \vspace{-2pt}
    \subsection{Setup}
    \label{sec:setup-tinystories}
    \vspace{-2pt}

    We selected four GPT-2-style Transformer baselines, covering varying numbers
    of layers and attention heads, including lightweight models (5M parameters) and
    medium-sized models (24M parameters). These models represent common
    attention-based architectures for short-context language modeling.

    Due to LuminaNet’s reliance on explicit connections and a cluster count that
    scales one-to-one with sequence length, it is inherently at a severe disadvantage
    in terms of parameter count and computational efficiency when compared to models
    optimized for efficient attention computation.

    To reduce parameter overhead, we only replaced the linear layers in the
    Input Layer of each cluster with a single shared embedding head
    \cite{chen2020simclr}. The network was fixed to 32 clusters, with each
    cluster processing one token; token-level dependencies were integrated through
    connections within the Neuron Layer. The model was trained for one epoch on
    the Nvidia A800 using the next-token prediction objective. Inference was performed
    on Snapdragon 8 Elite (mobile phone), Apple M2 (CPU), and Nvidia A800 (GPU).
    Training employed the AdamW optimizer with learning rate 0.001, weight decay
    0.1, $\beta_{1}= 0 .9$, $\beta_{2}= 0.95$, batch size 12,800, and a vocabulary
    size of 4,096.

    \subsection{Baseline Comparison}
    \begin{table}[]
        \scriptsize
        \centering
        \setlength{\tabcolsep}{3pt}
        \caption{Baseline comparison for the text generation task. Networks with
        a subscript "Conn" were initialized with a large number of pre-established
        connections. H refers to head and L denotes layer.}
        \vspace{5pt}
        \label{tab:baseline-tinystories}
        \begin{tabular}{@{}lccccr@{\ $\rightarrow$\ }lcccccc@{}}
            \toprule Model                          & Dim. & Pos.         & Params (M) & Structure   & \multicolumn{2}{c}{Evolve (Params)} & PPL    & Top-1 & Top-3 & Top-5 & FLOPs (T) & Mem. (G) \\
            \midrule $\text{GPT-2}_{\text{1H-12L}}$ & 384  & $\checkmark$ & 24.46      & Transformer & \multicolumn{2}{c}{-}               & 4.08   & 64.39 & 81.82 & 87.08 & 4.13      & 23.13    \\
            $\text{GPT-2}_{\text{12H-12L}}$         & 384  & $\checkmark$ & 24.46      & Transformer & \multicolumn{2}{c}{-}               & 4.02   & 64.66 & 82.07 & 87.29 & 4.13      & 23.13    \\
            $\text{GPT-2}_{\text{1H-1L}}$           & 384  & $\checkmark$ & 4.94       & Transformer & \multicolumn{2}{c}{-}               & 9.73   & 50.53 & 67.54 & 74.11 & 2.26      & 9.53     \\
            $\text{GPT-2}_{\text{12H-1L}}$          & 384  & $\checkmark$ & 4.94       & Transformer & \multicolumn{2}{c}{-}               & 8.08   & 53.29 & 70.59 & 77.02 & 2.27      & 9.65     \\
            $\text{LuminaNet}_{\text{Conn}}$        & 384  & -            & 43.24      & BNN         & 82.47M                              & 43.24M & 8.40  & 53.38 & 70.18 & 76.34     & 1.68    & 5.12 \\
            LuminaNet                               & 384  & -            & 44.37      & BNN         & 15.76M                              & 44.37M & 8.70  & 52.88 & 69.62 & 75.83     & 1.50    & 4.74 \\
            $\text{LuminaNet}_{\text{Conn}}$        & 256  & -            & 29.12      & BNN         & 36.13M                              & 29.12M & 11.22 & 49.12 & 65.35 & 71.72     & 1.09    & 3.85 \\
            LuminaNet                               & 256  & -            & 15.55      & BNN         & 6.31M                               & 15.55M & 13.49 & 47.31 & 63.20 & 69.51     & 0.55    & 2.53 \\
            \bottomrule
        \end{tabular}
    \end{table}
    As shown in Table~\ref{tab:baseline-tinystories}, LuminaNet achieves performance
    that is broadly comparable in terms of Top-k accuracy and perplexity, while
    reducing computational cost by approximately 25\% and peak memory consumption
    by nearly 50\% compared to single-layer GPT-2. This result demonstrates that
    even without attention or positional encoding, LuminaNet’s complex topological
    structure, self-evolved through architectural dynamics, is capable of capturing
    a certain degree of sequential dependencies.

    As expected, larger Transformers significantly outperform LuminaNet in
    generation quality, thanks to their self-attention architecture and stronger
    inductive biases for sequential modeling. However, this performance gain comes
    at the cost of substantially increased computational and memory overhead:
    their FLOPs and peak memory usage are orders of magnitude higher than those
    of LuminaNet. Although performance gaps persist after the same number of training
    epochs, this disparity largely stems from the fundamental paradigm
    difference between static and dynamic architectures.

    We will provide detailed quality assessments of the generated texts by
    LuminaNet in Appendix~\ref{supp:sentence-completion} -
    \ref{supp:story-generation}.

    \subsection{Ablation}
    \begin{table}[]
        \centering
        \scriptsize
        \setlength{\tabcolsep}{3pt}
        \caption{Performance differences of LuminaNet on TinyStories after
        removing connections.}
        \vspace{5pt}
        \label{tab:ablation-tinystories}
        \begin{tabular}{@{}lc|ccccc|ccccc@{}}
            \toprule                                  &      & \multicolumn{5}{c|}{Raw} & \multicolumn{5}{c}{Without Connections} \\
            \midrule Model                            & Dim. & Connections              & Params (M)                             & PPL   & Top-1 & FLOPs (T) & Params (M) & PPL                       & Top-1 & FLOPs (T) & Perf. Gap \\
            \midrule $\text{LuminaNet}_{\text{Conn}}$ & 384  & 149                      & 43.24                                  & 8.40  & 53.38 & 1.68      & 21.27      & 1$\times$10$^{\text{10}}$ & 0.41  & 0.59      & -99.23    \\
            LuminaNet                                 & 384  & 178                      & 44.37                                  & 8.70  & 52.88 & 1.50      & 18.12      & 52.04                     & 41.62 & 0.56      & -21.29    \\
            $\text{LuminaNet}_{\text{Conn}}$          & 256  & 316                      & 29.12                                  & 11.22 & 49.12 & 1.09      & 8.41       & 3$\times$10$^{\text{5}}$  & 5.03  & 0.18      & -89.76    \\
            LuminaNet                                 & 256  & 125                      & 15.55                                  & 13.49 & 47.31 & 0.55      & 7.36       & 62.22                     & 34.89 & 0.16      & -26.25    \\
            \bottomrule
        \end{tabular}
    \end{table}

    Table~\ref{tab:ablation-tinystories} reports the results of the connection ablation
    experiments. For networks that evolved from scratch without any initial
    connections, even after pruning all connections, performance does not
    decline sharply. This is because each cluster was initially trained independently
    and learned how to derive the appropriate token from the raw embedding vector.

    In contrast, for networks that began with a large number of pre-established
    connections, removing all connections causes immediate collapse. This is
    because, from the outset, all clusters could access sequence information
    through connections. Evolution during training served to eliminate redundant
    connections and establish new, useful ones. After pruning, each cluster can
    only receive the embedding of a single position, rendering it incapable of generating
    meaningful sequences.

    \subsection{Structure Visualization}
    \begin{figure*}
        \centering
        \includegraphics[width=1.0\textwidth]{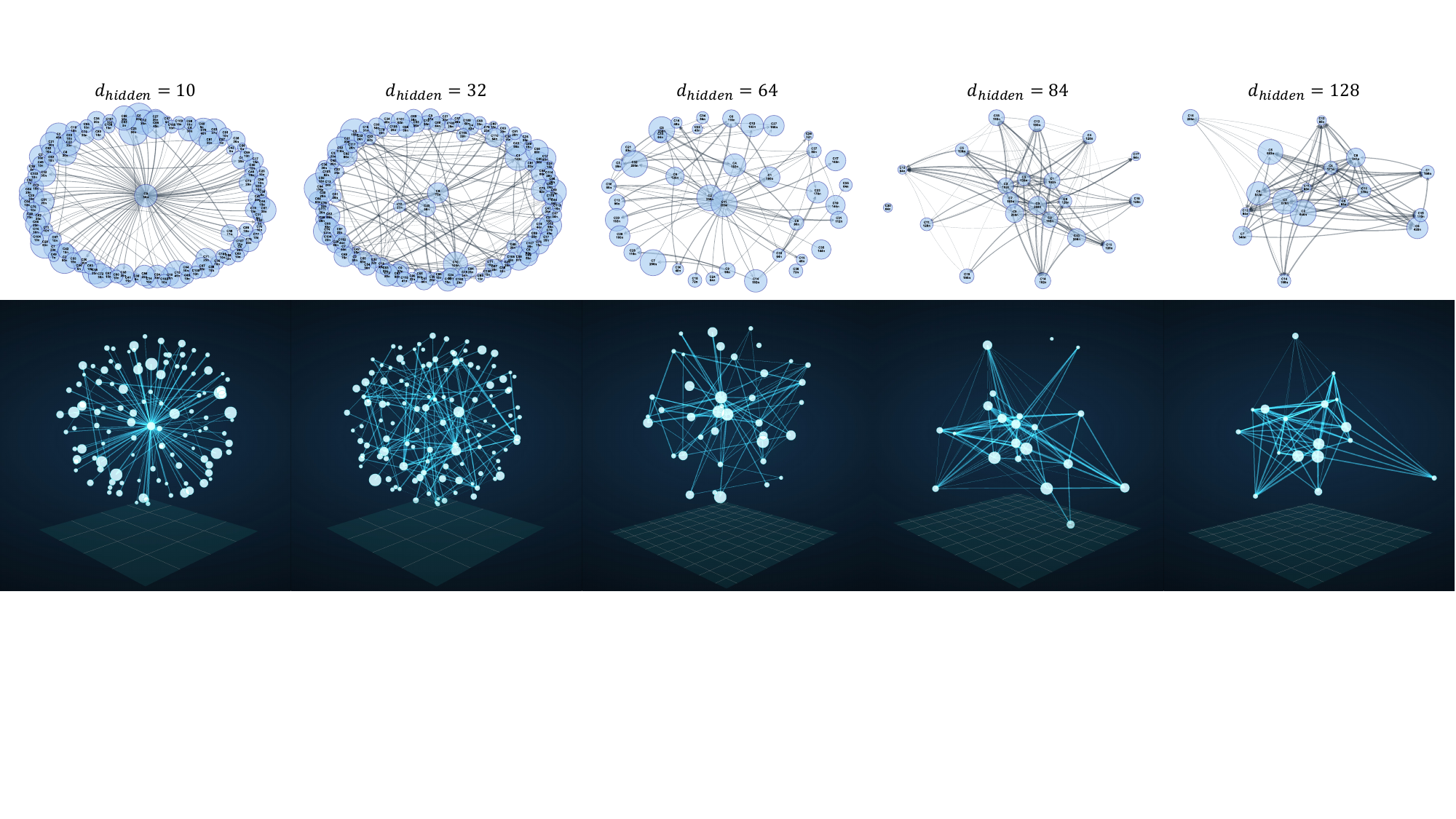}
        \caption{Architectures autonomously formed by LuminaNet on the
        TinyStories text generation task.}
        \label{fig:structure-tinystories}
    \end{figure*}

    Figure~\ref{fig:structure-tinystories} shows the network architectures
    formed on TinyStories task. Since the splitting strategy is disabled, the
    network is forced to enhance its information processing capacity through
    growth and connection formation. As the text generation task heavily relies on
    semantic associations between words, the resulting topological complexity
    far exceeds that observed in image recognition tasks.

    For networks with $d_{\text{hidden}}$ of 128 and 256, due to the relatively weaker
    capacity of individual clusters, the network requires more information from
    other clusters, ultimately forming topological depths of 21 layers as well as
    13 and 24 recurrent structures, respectively. In contrast, the network with
    $d_{\text{hidden}}$ of 384 achieves a shallower depth of 9 layers and only 4
    recurrent structures, as its individual clusters are more powerful and require
    less inter-cluster communication.

    Notably, two networks initialized with $d_{\text{hidden}}$ of 384 (one starting
    with pre-established dense connections and the other with no initial connections)
    converged to extremely similar topological architectures after evolution. This
    observation suggests that language sequences exhibit a relatively fixed pattern
    of semantic associations, which LuminaNet is able to capture even when
    evolving from completely opposite initialization directions.

    \vspace{-3pt}
    \section{Conclusion}
    \vspace{-3pt}
    \label{sec:conclusion}

    This paper rethinks the fundamental differences between current artificial neural
    networks and biological brain architecture, integrating insights from
    neuroscience to propose for the first time a novel paradigm of self-evolving
    artificial neural network, Brain-like Neural Network (BNN), and presents
    LuminaNet as a concrete implementation conforming to this paradigm.

    LuminaNet avoids convolutions, self-attention, positional encodings, and manual
    architectural design; it does not rely on prior assumptions, but instead evolves
    autonomously. On image recognition tasks, it surpasses classical convolutional
    architectures such as LeNet-5 and AlexNet, and outperforms several MLP/ViT
    models. Remarkably, it also achieves performance on par with single-layer
    Transformers in text generation tasks, demonstrating substantial potential.

    We provide the first empirical demonstration that artificial neural networks
    can autonomously construct and optimize themselves. Through extensive
    visualization experiments conducted on non-manually designed architectures, we
    comprehensively analyze the roles and operational mechanisms of different network
    modules, offering strong evidence for interpretability.

    We believe that the Brain-like Neural Networks will provide a novel
    perspective and approach for artificial intelligence, fostering deeper integration
    between AI and neuroscience.

    \newpage

    \bibliographystyle{unsrt}
    \bibliography{ref}

    \newpage
    \appendix

    \section{Related Work}
    \subsection{Brain Science}
    \textbf{Hebbian Learning.} Hebbian plasticity (“neurons that fire together
    wire together”) \cite{Hebb, bliss1973long, markram1997regulation, bi1998synaptic,bienenstock1982theory}
    constitutes a foundational principle of learning in biological neural networks.
    Recent studies have revisited Hebbian mechanisms in more biologically
    grounded settings. For instance, Eckmann et al. \cite{eckmann2024synapse} demonstrate
    through simulations that cortical networks can self-organize biologically plausible
    response properties under a synapse-type-specific competitive Hebbian learning
    rule acting jointly on excitatory and inhibitory synapses. Similarly, Halvagal
    et al. \cite{halvagal2023combination} show that combining Hebbian learning
    with predictive plasticity enables deep sensory networks to acquire
    invariant object representations, suggesting a close connection between classical
    Hebbian rules and modern representation learning.

    \textbf{Retinal Receptive Fields.} Retinal ganglion cells (RGCs)
    \cite{rgc,rgc-fund,rgc-meaning} typically exhibit center–surround receptive
    fields, a structure often interpreted under the efficient coding hypothesis \cite{karamanlis2025nonlinear}.
    Within the predictive coding framework, inhibitory surround circuits can be viewed
    as suppressing predictable components of central stimulation, thereby
    improving coding efficiency \cite{gupta2023panoramic}. Recent evidence further
    indicates that receptive field structures adapt to scene statistics across retinal
    locations. For example, Gupta et al. \cite{gupta2023panoramic} report
    systematic variations in receptive field shapes along the dorsal–ventral
    axis of the mouse retina, with asymmetric surround enhancement near the horizon,
    consistent with theoretical predictions. Moreover, under dynamic viewing
    conditions, Karamanlis et al. \cite{karamanlis2025nonlinear} show that
    correlations among RGCs deviate from pure decorrelation, suggesting that nonlinear
    pooling mechanisms preserve locally salient, high-contrast information
    during fixations \cite{karamanlis2025nonlinear}.

    \textbf{Lateral Geniculate Nucleus (LGN).} The lateral geniculate nucleus (LGN)
    \cite{lgn}, a thalamic relay between the retina and V1, preserves retinotopic
    organization and receives extensive cortical feedback \cite{ghodrati2017towards}.
    Although traditionally modeled as a largely linear relay, recent reviews emphasize
    its rich nonlinear properties, including dendritic nonlinearities, network synchronization
    and oscillations, and strong modulatory feedback from cortex
    \cite{ghodrati2017towards}. Experimental studies demonstrate that layer-6 cortical
    feedback can dynamically modulate LGN gain depending on visual context,
    selectively amplifying or suppressing responses to specific stimulus regions
    \cite{schmors2025effects}. Together, these findings suggest that the LGN
    actively participates in regulating visual information flow as part of the
    retina–thalamus–cortex loop, rather than serving as a passive relay \cite{ghodrati2017towards,
    schmors2025effects}.

    \textbf{V1 Structure and Hierarchical Organization.} Primary visual cortex (V1)
    \cite{retina-to-V1, lgn-to-v1} exhibits a canonical six-layer structure and,
    in some species, a columnar organization. In cats and primates, neurons are
    organized into orientation- and ocular-dominance columns spanning cortical layers,
    whereas rodents lack pronounced columnar architecture despite exhibiting
    strong orientation selectivity \cite{espinosa2012development}. Feedforward inputs
    from the LGN primarily target layer 4, with subsequent propagation to
    supragranular layers (layers 2/3) for feedforward projections to higher
    visual areas, while infragranular layers (layers 5/6) provide recurrent and feedback
    connections \cite{bastos2012canonical}. This laminar organization supports
    the classical simple-to-complex cell hierarchy within V1 and underlies
    progressive feature abstraction toward higher visual areas (V2, V4, IT). Notably,
    Federer et al. (2021) show that feedback projections from V2 to V1 preserve parallel
    channel organization consistent with feedforward pathways, indicating
    functionally specific feedback streams \cite{federer2021stream}.

    \textbf{Feedforward–Feedback Loops.} Feedforward and feedback connections are
    central to visual processing. Across the cortical hierarchy, feedforward pathways
    shape receptive fields in higher areas, while feedback pathways mediate
    attention, prediction, and contextual modulation \cite{federer2021stream}. Within
    V1 and beyond, these pathways form parallel ascending and descending streams.
    Along the entire visual pathway, the feedforward flow from retina to LGN to
    V1 interacts with corticothalamic feedback, enabling neural activity to integrate
    bottom-up sensory drive with top-down behavioral and state-dependent signals
    from cortex and brainstem.

    \subsection{Deep Learning}
    \textbf{Convolutional Neural Networks.} Convolutional Neural Networks (CNNs)
    incorporate local receptive fields and weight sharing, drawing inspiration from
    the structural organization of visual cortex, and achieve near-human
    performance in object recognition. A substantial body of work demonstrates strong
    correspondence between representations learned by hierarchical CNNs and neural
    responses in the primate visual system. In particular, higher CNN layers predict
    activity in inferotemporal (IT) cortex, while intermediate layers align with
    mid-level visual areas such as V4
    \cite{yamins2014performance, raman2020convolutional}. For example, Yamins et
    al. (2014) show that high-performing CNNs selected for object recognition accuracy
    exhibit strong correlations with IT neural responses, with intermediate
    layers predicting V4 activity. Subsequent reviews highlight striking similarities
    between CNNs and biological vision in both structure and function \cite{celeghin2023convolutional},
    while noting that standard CNNs largely omit parallel pathways and recurrent
    feedback ubiquitous in the brain. Recent efforts therefore explore augmenting
    CNNs with attention mechanisms and auxiliary visual pathways to better approximate
    biological vision \cite{celeghin2023convolutional}.

    \textbf{Attention Mechanisms.} Attention mechanisms in deep learning
    dynamically weight information by computing correlations among input
    features. Vaswani et al. \cite{transformer} introduce the Transformer
    architecture, replacing recurrence and convolution with self-attention based
    on query–key–value interactions, enabling highly parallel and efficient
    sequence modeling. Self-attention can be interpreted as assigning relevance scores
    across input representations, allowing models to emphasize informative
    components. Analogously, biological visual attention is mediated by
    frontoparietal networks that modulate sensory processing. Saliency-map-based
    modules commonly used in CNNs provide a computational analogue, guiding
    networks to focus on task-relevant image regions
    \cite{celeghin2023convolutional}. While the implementation details differ,
    artificial and biological attention share a functional role in selective
    information routing, improving robustness in complex scenes and multi-task settings,
    and informing models of selective information flow in cognition.

    \subsection{Broad Learning}
    An alternative to deep architectures is to increase network width to
    simplify training while maintaining expressive power. Chen et al.
    \cite{chen2017broad} propose the Broad Learning System (BLS), a flat network
    architecture that maps inputs to feature nodes and incrementally adds enhancement
    nodes without retraining the entire model, achieving efficient performance
    across various tasks \cite{chen2017broad}. More recent work by Errica et al.
    \cite{errica2025adaptive} introduces adaptive-width networks that jointly
    optimize neuron width and weights during training, allowing networks to expand
    or compress width based on task complexity. These models demonstrate width
    adaptability across tabular, image, and sequential data, and enable post-training
    pruning to balance performance and computational cost \cite{errica2025adaptive}.
    Together, these studies represent early explorations into width-centric learning
    paradigms, offering complementary perspectives to depth-focused network
    design.

    \section{Evolution of the Input Layer Encoder}
    \label{supp:encoder-evolution}
    \begin{figure*}
        \centering
        \includegraphics[width=1.0\textwidth]{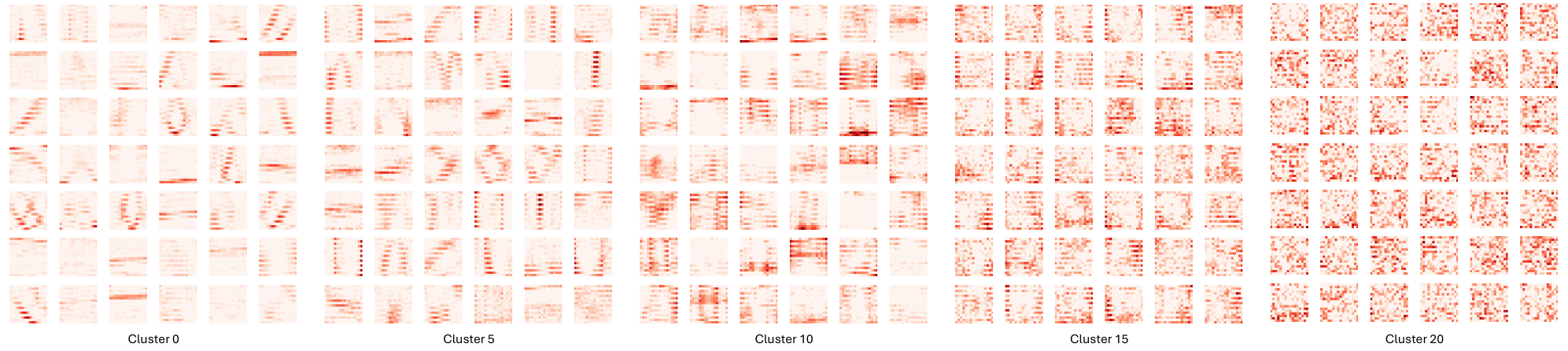}
        \caption{The evolutionary process of the encoder matrix of Neuron
        Clusters.}
        \label{fig:embedding-evolve}
    \end{figure*}
    Clusters with smaller indices have existed in the network for longer and
    undergone more training epochs. As shown in Figure~\ref{fig:embedding-evolve},
    the encoder weight matrices of the Input Layer for both new and old clusters
    are visualized, revealing that during evolution, cluster encoders transition
    from randomly initialized feature representations to developing distinct, specialized
    feature preference patterns.

    \section{Connection Weights}
    \label{supp:connection-weights}
    \begin{figure*}
        \centering
        \includegraphics[width=1.0\textwidth]{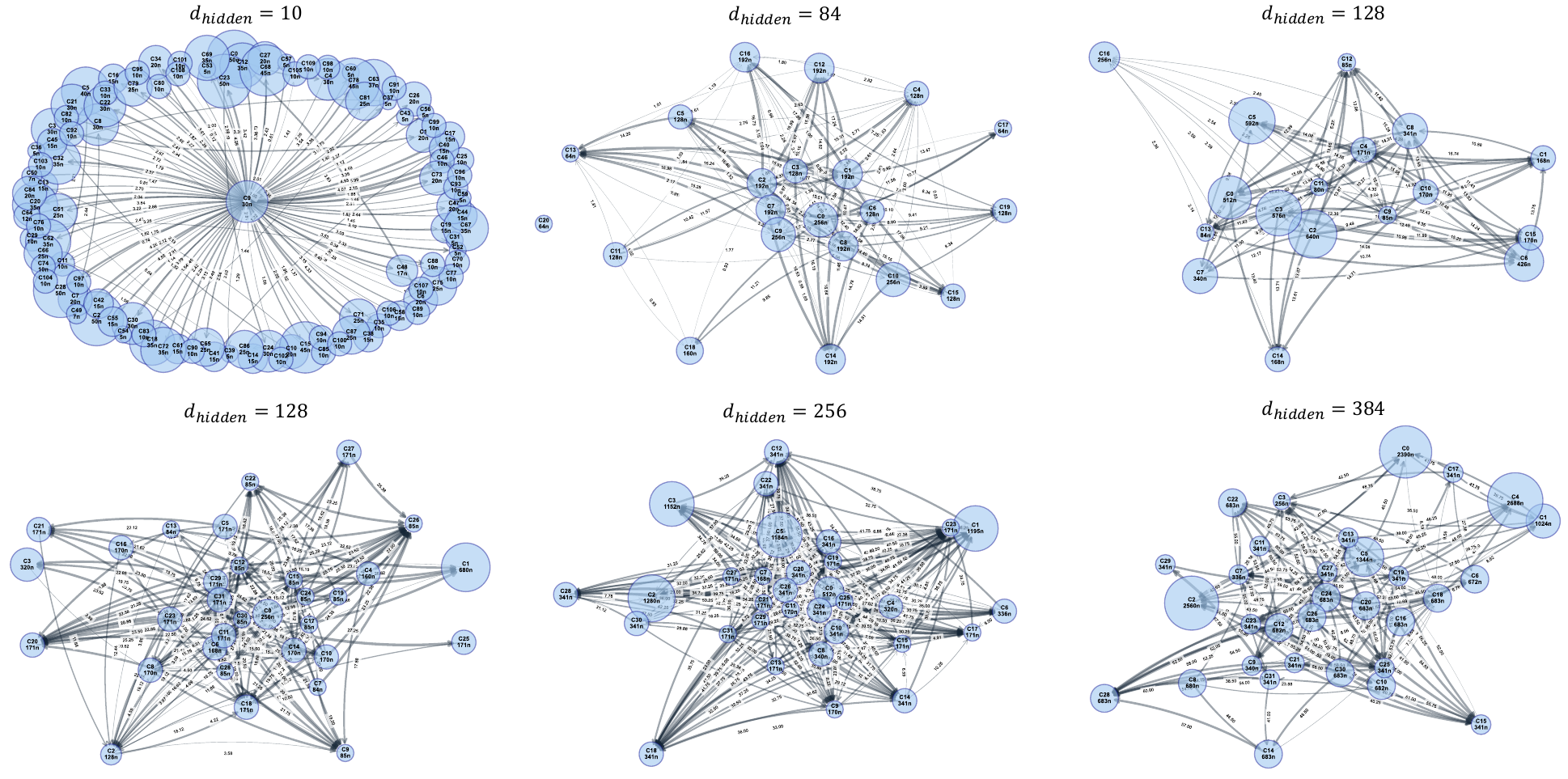}
        \caption{Quantitative analysis of the connection strengths and the
        architecture. The top row is the image recognition network, and the
        bottom row is the text generation network.}
        \label{fig:structure-label}
    \end{figure*}
    Following the connection strength definition method described in Section~\ref{sec:strategies},
    we quantitatively characterize the strength of each connection. As shown in Figure~\ref{fig:structure-label},
    connection strengths in the language generation task are significantly
    higher than those in the image recognition task. This indicates that connections
    in the former task enable more diverse transformations of the raw output signals
    from neurons.

    This phenomenon can be attributed to two factors: First, the disabling of the
    splitting strategy forces information exchange to rely solely on connections,
    thereby amplifying their functional importance. Second, language generation
    is intrinsically more challenging than image recognition: its discretized inputs
    and outputs, long-range semantic dependencies, and complex syntactic structures
    all demand stronger representational capacity from the network.

    Given that the base architecture of LuminaNet is deliberately simple and incorporates
    no manually injected inductive biases, the network can only acquire enhanced
    expressive power by continuously establishing and optimizing connections, thereby
    evolving increasingly complex topological structures.

    \section{Sentence Completion}
    \label{supp:sentence-completion}
    \begin{table}[]
        \centering
        \scriptsize
        \caption{Outputs of LuminaNet with varying scales and initializations on
        the story completion task.}
        \vspace{5pt}
        \label{tab:completion}
        \begin{tabularx}
            {\textwidth}{@{}X|p{2cm}p{2cm}p{2cm}p{2cm}@{}} \toprule Prompt &
            $\text{LuminaNet-384}_{\text{Conn}}$ & LuminaNet-384 &
            $\text{LuminaNet-256}_{\text{Conn}}$ & LuminaNet-256 \\
            \midrule Once upon a time , in an ancient house , there lived a girl
            named & Lily. & Sue. & Sally. & Amy. \\
            \midrule Lily likes to eat & potatoes. & her dinner. & yogurt. & . \\
            \midrule Jack’s mother was not home, and his father was at home. When
            Jack came home, he said hello to & his mum and dad. & him. & his mom
            and dad. & Jack. \\
            \midrule Jack wanted to read a book, so he went to & the store to buy
            a new one. & the library. & the park. & visit the store. \\
            \midrule Alice was so tired when she got back home so she went & to the
            kitchen to find something to eat. & off to sleep. & to the lake. &
            to bed. \\
            \midrule On weekends Jack went to visit his grandmother whereas on weekdays
            he would go to school. Last weekend, when Jack was on his way to & to
            his friend Jack' s house. & get the bus to go home. & help him. & school.
            \\
            \midrule If I throw a ball up in the air, eventually it will & fly around
            and around. & be so much fun! & not go to the park. & see it was so
            pretty. \\
            \midrule The pumpkin replied, ”I am sad because I cannot & play
            outside. & go to the store. & fly like a bird. & in a big. \\
            \midrule "What do birds like to eat?", Tom asked & for a big jar of
            honey. & . "I want to play with the other kids . & her mom if she
            could go to the park. & "What do birds like to eat?" \\
            \midrule Lily likes cats and dogs . She asked her mom for a dog and
            her mom said no , so instead she asked & the lady for her pet. & "Can
            I have a cake now?" & her mom if they could do it. & . \\
            \midrule It was soft and gentle, and it said & , "Look mum, we can
            go fishing." & no. & they were a little boy. & . \\
            \midrule "What are you doing up there, Lucy?" Lucy replied, & "I' m making
            cookies and I' m going to make my house clean!" & "I don' t know how
            to be nice to each other . Lucy was happy to have an idea." & "It ' s
            ok, it' s not a big, red apple. You can have it." & pointing up a big
            cloud in the sky. \\
            \midrule First, she had to climb to a nearby rock, then she had to
            crawl to the cliff and then herassador and back. Finally, she
            managed to reach & the top of the tree. & the top of the hill. & the
            bird, and the little girl was no longer scared. & her the biggest ,
            she could be safe. \\
            \midrule Jack and Lily saw a rainbow after a rainy day. They were
            amazed by the colors. Jack said, "Look, Lily. A rainbow has & a picture
            of our wish. & a pink dress and a rainbow. & a picture of a princess!"
            & a nice in the park. \\
            \midrule It was winter and cold outside so his mother told him, "You
            should & always listen to me and be careful when you play in the
            snow." & have a lot of fun. & not get out. & always go outside in
            the sunshine, but you can always be dangerous." \\
            \bottomrule
        \end{tabularx}
    \end{table}

    We adopt the commonly used short-sentence completion evaluation protocol
    from the TinyStories dataset \cite{TinyStories} to assess the network’s
    capability in local semantic understanding and text continuation. Specifically,
    we truncate several story sentences as input prompts, and require the network
    to generate subsequent short sentences without relying on additional context.
    Table~\ref{tab:completion} summarizes the performance of different network
    configurations on this task.

    From the results, we observe that all networks are generally capable of generating
    grammatically plausible and reasonably coherent text in most cases,
    demonstrating good consistency in naming entities, locations, and common
    behavioral descriptions. However, clear differences remain among
    architectures in terms of semantic coherence and completion quality. Networks
    initialized with pre-established connections tend to produce more specific and
    semantically grounded completions, often featuring more explicit verbs or
    objects and exhibiting stronger contextual continuity in behavioral
    descriptions and scenario extensions. In contrast, networks initialized
    without connections are more prone to generating conservative or prematurely
    terminated outputs, with some examples producing only punctuation marks or semantically
    incomplete phrases.

    Additionally, we note that the networks do not simply reproduce high-frequency
    templates from the training set; instead, they generate diverse
    continuations tailored to different prompts. This indicates that even without
    attention mechanisms or positional encoding, the proposed architecture
    contains a certain degree of local language modeling capability.

    \section{Long text generation}
    \label{supp:story-generation}

    We further evaluate LuminaNet’s long-text generation capability on the TinyStories
    dataset. Following the text quality evaluation paradigm established in the
    original TinyStories work \cite{TinyStories}, we use GPT-5.2 \cite{openai_gpt5}
    to assess the generated text across three dimensions: Grammar (grammatical correctness),
    Creativity (originality and stylistic variation), and Consistency (semantic coherence
    and narrative continuity).

    We use the following prompt to instruct GPT-5.2:

    \begin{small}
        \texttt{You are a primary school teacher and are grading the short
        stories written by the students. Please rate the stories based on 3 criteria:
        grammar, creativity, and consistency. The score ranges from 1 to 10. Please
        output in the following format: \\
        Grammar: <score>/10 \\
        Creativity: <score>/10 \\
        Consistency: <score>/10}
    \end{small}

    \begin{table}[h]
        \centering
        \scriptsize
        \caption{The stories generated by the network with $d_{\text{hidden}}$
        of 384 obtained through different initialization methods and their
        quality evaluations.}
        \vspace{5pt}
        \label{tab:story1}
        \begin{tabularx}
            {\textwidth}{@{}lX|l@{}} \toprule Model & Generated Text & Score \\
            \midrule \multirow{4}{*}{$\text{LuminaNet-384}_{\text{Conn}}$} & Anna
            and Ben are friends. They like to play with their toys in their room.
            But today, they have a big box of blocks in their room. They have to
            put it on their shelf. They have fun. They are happy. &
            \begin{tabular}[t]{@{}l@{}}
                Grammar: 8/10     \\
                Creativity: 4/10  \\
                Consistency: 8/10
            \end{tabular}
            \\
            \cmidrule(l){2-3} & Once upon a time, there was a little girl named Lily.
            She loved to play outside in the forest. One day, she found a big
            flower on the ground. She picked it up and showed it to her mom. "Mommy,
            look at all the flowers" she exclaimed. Her mom smiled and said, "That'
            s a special gift. It' s so beautiful, I love it." &
            \begin{tabular}[t]{@{}l@{}}
                Grammar: 8/10     \\
                Creativity: 6/10  \\
                Consistency: 7/10
            \end{tabular}
            \\
            \cmidrule(l){2-3} & Once upon a time, there was a little girl named Lily.
            She loved to play outside in the sunshine. One day, she saw a big, shiny
            ball on the ground. She picked it up and took it home. The little girl
            was so happy and said, "Wow, this is the best day ever." &
            \begin{tabular}[t]{@{}l@{}}
                Grammar: 9/10     \\
                Creativity: 5/10  \\
                Consistency: 9/10
            \end{tabular}
            \\
            \cmidrule(l){2-3} & Once upon a time there was a girl named Sarah. She
            was three years old and loved playing hide and seek. One day, she found
            a big, green ball. She was so excited that she wanted to explore it.
            She asked her mommy, "Can I swing in the rain, please?" Her mom said,
            "Yes, of course you can." So they went to the park, Lily and her friends
            went to the park together. They played tag and had lots of fun. When
            it was time to go home, Lily was tired but happy. She had been so happy
            and excited to have her new toy. She hugged her mom and dad and said,
            "Thank you for playing with me, my dear. You are my best friend." &
            \begin{tabular}[t]{@{}l@{}}
                Grammar: 7/10     \\
                Creativity: 6/10  \\
                Consistency: 6/10
            \end{tabular}
            \\
            \midrule \multirow{4}{*}{LuminaNet-384} & Once upon a time, there
            was a little girl named Lily. She loved to play outside in the sunshine.
            One day, she was playing with her friends when she saw a beautiful rainbow.
            It was a very big, so she picked it up and showed it to her mom. "Let'
            s see what it in the sky," she said. So, she put it in the envelope
            and opened it. She saw a big bag of jewel s that she was so happy.
            She picked it up and showed it to her mom and dad. She had a great day
            at the park and the girl was so proud of her. She wanted to make
            something amazing. She went to her bedroom and saw the animals. She
            thought it was funny. She decided to take a step on a trip. The girl
            was so excited. She asked him if she wanted to take the closer look.
            Then, they started walking along the river. The girl was very
            excited and wanted to join her. She went to the pond and found a small
            fish. She picked up the water and put it in the water. The water was
            so happy to see her mom. She felt proud of herself. She said, "I will
            be so good! I can play with the bubbles again tomorrow. &
            \begin{tabular}[t]{@{}l@{}}
                Grammar: 5/10     \\
                Creativity: 8/10  \\
                Consistency: 4/10
            \end{tabular}
            \\
            \cmidrule(l){2-3} & Once upon a time there was a mighty bear. He was
            very big and strong and he was going on a trip. One day he found something
            very special. It was a big and scary bear! The bear was afraid. He
            wanted to see his friend. He ran around the door and opened it.
            Inside, there was a big bear with a bear and bear. He was scared and
            didn' t know what to do. But then, he heard a voice from behind him.
            It was a big bear. "What are you doing here?" The little boy asked, "I
            am trying to find a way to get the water. I' m so happy." The little
            girl thanked the little bird for the help. From then on, she was
            careful and the girl became good friends. She helped me carry the twig
            s and the bird, and the girl always remembered to be careful. &
            \begin{tabular}[t]{@{}l@{}}
                Grammar: 4/10     \\
                Creativity: 7/10  \\
                Consistency: 3/10
            \end{tabular}
            \\
            \cmidrule(l){2-3} & Once upon a time there was a little girl who loved
            to play in the garden. One day, she found a little girl called Daisy
            and the lizard. She was so happy and excited to help her mom. The little
            girl said "Hello, I' m glad you like it. She gave me a big hug. The
            little girl smiled and hugged her mom. She was happy she had a friend.
            &
            \begin{tabular}[t]{@{}l@{}}
                Grammar: 6/10     \\
                Creativity: 5/10  \\
                Consistency: 6/10
            \end{tabular}
            \\
            \cmidrule(l){2-3} & Once upon a time, there was a big, hairy dog
            named Spot. Spot loved to run and play outside. One day, Spot was playing
            with a ball when he saw a big, round ball. He was so excited and wanted
            to play with it. He ran to the ball with his ball and a ball. He kicked
            it. He threw it high and the ball rolled high in the air. He ran and
            ran, but his mom told him it. "I want to get the bird" The boy got angry.
            He flew to the boy' s mom gave him a big hug. The boy said, "Don' t worry,
            I' ll help you get a good boy." Timmy smiled and his mom went to the
            kitchen. After the boy' s mom gave him some milk and he took out to the
            sink. When he got to the store, he saw the boy was very excited and wanted
            to try it. He was so excited and he started to get the toy. But the little
            boy was so happy to have found the toy. The little girl was so happy
            to have the little girl. She thanked the little girl for her kindness
            and played together. They all had fun. &
            \begin{tabular}[t]{@{}l@{}}
                Grammar: 4/10     \\
                Creativity: 6/10  \\
                Consistency: 3/10
            \end{tabular}
            \\
            \bottomrule
        \end{tabularx}
    \end{table}

    \begin{table}[h]
        \centering
        \scriptsize
        \caption{The stories generated by the network with $d_{\text{hidden}}$
        of 256 obtained through different initialization methods and their
        quality evaluations.}
        \vspace{5pt}
        \label{tab:story2}
        \begin{tabularx}
            {\textwidth}{@{}lX|l@{}} \toprule Model & Generated Text & Score \\
            \midrule \multirow{4}{*}{$\text{LuminaNet-256}_{\text{Conn}}$} & Once
            upon a time, there was a girl named Lily. Lily was very sad because she
            loved to play with her toys. One day, Lily' s mommy gave her a big
            bag of cheese. Lily was very excited to see her mommy. They walked
            and walked, but the car was too heavy. Lily was so happy, she hugged
            it tightly and it made her feel happy. &
            \begin{tabular}[t]{@{}l@{}}
                Grammar: 6/10     \\
                Creativity: 6/10  \\
                Consistency: 6/10
            \end{tabular}
            \\
            \cmidrule(l){2-3} & Once upon a time, there was a boy named Timmy.
            Timmy loved to play with his toys and run around outside. One day,
            Timmy and his friends went to play with his friends. They played on the
            swings and the slide, but Timmy didn' t know what to do. The end. &
            \begin{tabular}[t]{@{}l@{}}
                Grammar: 7/10     \\
                Creativity: 5/10  \\
                Consistency: 7/10
            \end{tabular}
            \\
            \cmidrule(l){2-3} & Once upon a time, there was a little boy named
            Timmy. He loved to play with his toys and run around outside. One
            day, Timmy and his friends played on the swings. They saw a big rock
            with a long tail. Tim was very happy and said, "Wow, Timmy! I want to
            play with you." They walked and walked until they found the way home.
            They saw many animals and toys to go on it. The car was happy. But
            then, something bad happened. It was the best day ever! &
            \begin{tabular}[t]{@{}l@{}}
                Grammar: 6/10     \\
                Creativity: 6/10  \\
                Consistency: 5/10
            \end{tabular}
            \\
            \cmidrule(l){2-3} & Once upon a time there was a little boy named Tim.
            Tim was very excited because he was feeling very tired. He loved to play
            with his friends and make new friends. One sunny day, Tim and his friends
            decided to play hide and seek, and he had a great time playing with
            the blocks in the world. &
            \begin{tabular}[t]{@{}l@{}}
                Grammar: 6/10     \\
                Creativity: 5/10  \\
                Consistency: 6/10
            \end{tabular}
            \\
            \midrule \multirow{4}{*}{LuminaNet-256} & Once upon a time, there was
            a little girl named Lily. She had a big, fluffy she loved to play with
            her toys. One day, her mom took her to the park to play. They were
            very excited and had lots of fun. But then on a big, Lily saw a big
            dog in the park. The dog wanted to see it was too. Lily ran to the
            fence. It was a big dog and the dog. The dog barked and barked
            barked. The dog barked barking and barked its bark. Lily and Tom
            were scared. Their mom wanted to help them. She saw them too and the
            dog. She told them to stop and the dog. She gave them a big hug and
            said, " I love you, Ben. And you have to be careful and share to play
            with us. I will not to be nice again." Ben says. " Okay, Mom. I promise.
            You will be good friends. I will be careful." Lily and Ben smile. &
            \begin{tabular}[t]{@{}l@{}}
                Grammar: 3/10     \\
                Creativity: 6/10  \\
                Consistency: 3/10
            \end{tabular}
            \\
            \cmidrule(l){2-3} & Once upon a time, there was a little girl named Lily.
            She loved to play in the park with her friends. One day, Lily' s friends
            told her they were going to the park. Lily wanted to play with her
            toys. Lily was so happy to see the sun rise. She saw a big slide
            with the slide and it down. She said, "This is my birthday. I can go
            to the zoo and I want to play with you," Lily said. She was sad. She
            said, "I want to play with you. It was fun too. And you can play
            with me. Do you want to play with me?" Lily said. "Yes, please, I want.
            I like a new doll. I like a big doll!" Tom said. He was happy. He said,
            "No, Sara' s doll. I love it is too. They were not good. They liked
            to play. They have to share and toys. They are sorry. &
            \begin{tabular}[t]{@{}l@{}}
                Grammar: 4/10     \\
                Creativity: 6/10  \\
                Consistency: 4/10
            \end{tabular}
            \\
            \cmidrule(l){2-3} & Once upon a time there was a little girl named Mia.
            She was three years old and loved to explore. One day she asked her dad,
            " Can I go on the door ?" Her mom said, " Yes, we can find a place to
            be a special place to the zoo." They walked inside and saw a big,
            green tree. She thought it was so pretty and she wanted to see it.
            She asked her mom, " Can I can see the butterfly' s mom ?" Her mom replied,
            " Yes, but can you have a new butterfly. It is a very special and
            you can fly to be careful. Do you want to play with it ?" Lily
            nodded and said, " Yes, I do. I want to try a new. But then, I have something
            else to do. Do you want to play with me ?" Lily says. She looks at
            the doll. He says, " It looks a good idea. I can play with my dolls.
            It is blue. You can do anything. He likes it." Lily and Ben smile. He
            says. " Thank you, Ben. You are nice. I will be good friend." They hug
            the car. They play on the swings. They have fun. They are friends. &
            \begin{tabular}[t]{@{}l@{}}
                Grammar: 4/10     \\
                Creativity: 7/10  \\
                Consistency: 4/10
            \end{tabular}
            \\
            \cmidrule(l){2-3} & Once upon a time, there was a little girl named
            Lily. She loved to sing in the sun with her friends. One day, her mom
            told her that she had a big surprise that her. She was not to her
            mommy so she went to the doctor. Her dad told her that was important
            to take her and said, but sometimes it was too late. The little girl
            was very happy because she was able to help her mommy and the girl.
            She was happy to help her mom and dad her. &
            \begin{tabular}[t]{@{}l@{}}
                Grammar: 3/10     \\
                Creativity: 5/10  \\
                Consistency: 3/10
            \end{tabular}
            \\
            \bottomrule
        \end{tabularx}
    \end{table}

    Table~\ref{tab:story1} - \ref{tab:story2} presents representative generation
    samples and corresponding scores for different network configurations. From
    the evaluation results, we observe clear differences across the three dimensions
    of text quality among the networks. Overall, networks initialized with pre-established
    connections tend to produce higher semantic consistency and structural integrity,
    with generated text typically unfolding around a single context without
    abrupt jumps or irrelevant insertions. This is because each cluster can initially
    access information from multiple sequence positions, enabling more effective
    modeling of semantic dependencies between words. In contrast, networks
    initialized without connections, due to their initial inability to communicate
    information across positions, often exhibit theme drift, repetition, or
    premature termination during generation.

    In terms of grammar, all networks generate text that is generally readable,
    but some networks exhibit word-order confusion, ambiguous pronoun references,
    or lexical repetition during long-distance generation, leading to lower grammar
    scores. Creativity scores reflect stylistic differences among the networks: some
    generate structurally simple yet stable narratives, while others produce
    more varied content, often at the cost of consistency.

    In the later stages of generation, we observe a significant decline in
    content quality, manifested as semantic drift, narrative breaks, and sudden insertions
    of irrelevant content. We attribute this phenomenon to the absence of memory
    mechanisms: as the generation progresses, the network cannot retain early
    context, leading to gradual forgetting of the established narrative.
    Furthermore, local prediction errors accumulate over time, amplifying
    semantic inconsistency and ultimately manifesting as semantic collapse in
    the later portion of the output.

    Nevertheless, even without attention mechanisms, positional encoding, causal
    masking, or memory modules, LuminaNet is capable of generating complete,
    multi-sentence story texts, demonstrating its inherent ability to model and organize
    long-term content. These results further validate our key insight: \textbf{the
    key to self-evolution in artificial neural networks lies in the complex
    topological structures formed by neuron connections.}

    \section{Discussion}
    \label{supp:discussion}

    To demonstrate the feasibility of self-evolving Brain-like Neuron Network
    paradigms, LuminaNet deliberately avoids most of the sophisticated architectures
    and optimization techniques developed in modern deep learning. Instead, it relies
    solely on basic multilayer perceptron (MLP) and standard gradient descent,
    with the goal of exploring whether a genuine paradigm shift at the
    architectural level is possible. This design choice directly results in lower
    computational efficiency compared to existing mainstream static neuron network
    architectures, while the dynamic architectural reconstructions during
    runtime further lead to a noticeable reduction in optimization efficiency.

    Nevertheless, these limitations are not fundamental obstacles. Rather, they stem
    from the fact that self-evolving dynamic neural network paradigms remain at an
    early stage, lacking a mature optimization ecosystem. With continued exploration
    and development in this area, these issues are expected to be addressed and
    mitigated at the technical level.

    \section{Future Work}
    \label{supp:future-work}

    Although LuminaNet demonstrates the feasibility of self-evolving BNNs, the
    proposed framework is still at an early stage and leaves substantial room
    for further exploration and improvement. Future work will focus on advancing
    both the computational efficiency and the evolutionary intelligence of the network,
    as well as extending its scope toward more ambitious and biologically
    grounded directions. Specifically, we outline several promising research
    avenues as follows:

    \begin{enumerate}[leftmargin=*, labelsep=1em]
        \item \textbf{Parallelization of Forward Propagation.} The forward
            propagation in LuminaNet is currently executed in a sequential
            manner. However, once dependencies are properly resolved, the Two-Pass
            Forward mechanism can naturally support parallel computation, which
            has the potential to significantly improve inference efficiency.

        \item \textbf{Adaptive Evolution Strategies.} At present, strategy selection
            during evolution in LuminaNet relies on hard-coded probability values.
            Enabling the network to autonomously learn and adjust the
            probabilities of different evolutionary strategies would make the
            evolution process more flexible and adaptive.

        \item \textbf{Autonomous Evolution Decision.} The timing of evolution in
            LuminaNet is currently determined by manually designed decision
            rules. Allowing the network to autonomously decide when to evolve would
            substantially advance the development of self-evolving Brain-like Neuron
            Networks.

        \item \textbf{Evaluation of Evolutionary Quality.} Not every
            evolutionary step is necessarily beneficial. Developing principled methods
            to evaluate whether a given evolution improves the model, and to
            quantify the relative quality of the network after evolution, is of fundamental
            importance for establishing reliable benchmarks for self-evolving BNNs.

        \item \textbf{Internet-Scale Distributed Artificial Neural Networks.}
            Extensive experiments have demonstrated that LuminaNet is capable of
            dynamic construction, dynamic reconfiguration, and dynamic inference.
            Building on this foundation, future work may further extend the
            framework following the philosophy of BNNs, by replacing Neuron Clusters
            with more powerful functional units and distributing them across arbitrary
            nodes on the Internet. New neuron nodes could be seamlessly
            integrated into the global network, while connections could be implemented
            as real-time network requests routed to appropriate nodes for inference
            and computation. Routing policies could be dynamically updated to
            optimize overall system performance and to identify nodes that are
            better suited for specific computational tasks.

        \item \textbf{Interdisciplinary Integration of Artificial Intelligence
            and Neuroscience.} The field of artificial intelligence is
            experiencing explosive technological growth, with large-scale foundation
            models and their industrial deployments continuously pushing humanity
            closer to the threshold of artificial general intelligence (AGI). We
            believe that achieving AGI will inevitably require AI systems to
            increasingly emulate the structure, dynamics, and developmental
            processes of the biological brain. In this work, we propose Brain-like
            Neuron Networks as an initial step toward this goal. Future research
            will further strengthen the interaction between AI and neuroscience,
            incorporating insights from brain connectivity, neural plasticity,
            and developmental mechanisms to guide the design of more
            biologically grounded, adaptive, and generalizable intelligent
            systems.
    \end{enumerate}


    \newpage
    \section*{NeurIPS Paper Checklist}

    \begin{itemize}
        \item {\bf Claims}

        \item[] Question: Do the main claims made in the abstract and
            introduction accurately reflect the paper's contributions and scope?

        \item[] Answer: \answerYes{} 

        \item[] Justification: We provide clear contributions and scope of the
            paper in Section~\ref{sec:intro}.

        \item[] Guidelines:
            \begin{itemize}
                \item The answer NA means that the abstract and introduction do not
                    include the claims made in the paper.

                \item The abstract and/or introduction should clearly state the claims
                    made, including the contributions made in the paper and
                    important assumptions and limitations. A No or NA answer to this
                    question will not be perceived well by the reviewers.

                \item The claims made should match theoretical and experimental results,
                    and reflect how much the results can be expected to
                    generalize to other settings.

                \item It is fine to include aspirational goals as motivation as long
                    as it is clear that these goals are not attained by the paper.
            \end{itemize}

        \item {\bf Limitations}

        \item[] Question: Does the paper discuss the limitations of the work
            performed by the authors?

        \item[] Answer: \answerYes{} 

        \item[] Justification: We provide discussion of limitations in Appendix~\ref{supp:discussion}.

        \item[] Guidelines:
            \begin{itemize}
                \item The answer NA means that the paper has no limitation while
                    the answer No means that the paper has limitations, but
                    those are not discussed in the paper.

                \item The authors are encouraged to create a separate "Limitations"
                    section in their paper.

                \item The paper should point out any strong assumptions and how robust
                    the results are to violations of these assumptions (e.g.,
                    independence assumptions, noiseless settings, model well-specification,
                    asymptotic approximations only holding locally). The authors
                    should reflect on how these assumptions might be violated in
                    practice and what the implications would be.

                \item The authors should reflect on the scope of the claims made,
                    e.g., if the approach was only tested on a few datasets or
                    with a few runs. In general, empirical results often depend
                    on implicit assumptions, which should be articulated.

                \item The authors should reflect on the factors that influence the
                    performance of the approach. For example, a facial recognition
                    algorithm may perform poorly when image resolution is low or
                    images are taken in low lighting. Or a speech-to-text system
                    might not be used reliably to provide closed captions for
                    online lectures because it fails to handle technical jargon.

                \item The authors should discuss the computational efficiency of
                    the proposed algorithms and how they scale with dataset size.

                \item If applicable, the authors should discuss possible
                    limitations of their approach to address problems of privacy
                    and fairness.

                \item While the authors might fear that complete honesty about limitations
                    might be used by reviewers as grounds for rejection, a worse
                    outcome might be that reviewers discover limitations that
                    aren't acknowledged in the paper. The authors should use their
                    best judgment and recognize that individual actions in favor
                    of transparency play an important role in developing norms that
                    preserve the integrity of the community. Reviewers will be
                    specifically instructed to not penalize honesty concerning
                    limitations.
            \end{itemize}

        \item {\bf Theory assumptions and proofs}

        \item[] Question: For each theoretical result, does the paper provide the
            full set of assumptions and a complete (and correct) proof?

        \item[] Answer: \answerYes{} 

        \item[] Justification: We provide clear and complete theories.

        \item[] Guidelines:
            \begin{itemize}
                \item The answer NA means that the paper does not include theoretical
                    results.

                \item All the theorems, formulas, and proofs in the paper should
                    be numbered and cross-referenced.

                \item All assumptions should be clearly stated or referenced in the
                    statement of any theorems.

                \item The proofs can either appear in the main paper or the supplemental
                    material, but if they appear in the supplemental material, the
                    authors are encouraged to provide a short proof sketch to provide
                    intuition.

                \item Inversely, any informal proof provided in the core of the
                    paper should be complemented by formal proofs provided in
                    appendix or supplemental material.

                \item Theorems and Lemmas that the proof relies upon should be properly
                    referenced.
            \end{itemize}

        \item {\bf Experimental result reproducibility}

        \item[] Question: Does the paper fully disclose all the information
            needed to reproduce the main experimental results of the paper to
            the extent that it affects the main claims and/or conclusions of the
            paper (regardless of whether the code and data are provided or not)?

        \item[] Answer: \answerYes{} 

        \item[] Justification: The experimental results can be reproduced.

        \item[] Guidelines:
            \begin{itemize}
                \item The answer NA means that the paper does not include experiments.

                \item If the paper includes experiments, a No answer to this
                    question will not be perceived well by the reviewers: Making
                    the paper reproducible is important, regardless of whether
                    the code and data are provided or not.

                \item If the contribution is a dataset and/or model, the authors
                    should describe the steps taken to make their results
                    reproducible or verifiable.

                \item Depending on the contribution, reproducibility can be
                    accomplished in various ways. For example, if the
                    contribution is a novel architecture, describing the architecture
                    fully might suffice, or if the contribution is a specific
                    model and empirical evaluation, it may be necessary to either
                    make it possible for others to replicate the model with the same
                    dataset, or provide access to the model. In general.
                    releasing code and data is often one good way to accomplish
                    this, but reproducibility can also be provided via detailed instructions
                    for how to replicate the results, access to a hosted model (e.g.,
                    in the case of a large language model), releasing of a model
                    checkpoint, or other means that are appropriate to the
                    research performed.

                \item While NeurIPS does not require releasing code, the
                    conference does require all submissions to provide some
                    reasonable avenue for reproducibility, which may depend on the
                    nature of the contribution. For example
                    \begin{itemize}
                        \item If the contribution is primarily a new algorithm,
                            the paper should make it clear how to reproduce that
                            algorithm.

                        \item If the contribution is primarily a new model architecture,
                            the paper should describe the architecture clearly
                            and fully.

                        \item If the contribution is a new model (e.g., a large
                            language model), then there should either be a way
                            to access this model for reproducing the results or
                            a way to reproduce the model (e.g., with an open-source
                            dataset or instructions for how to construct the dataset).

                        \item We recognize that reproducibility may be tricky in
                            some cases, in which case authors are welcome to
                            describe the particular way they provide for
                            reproducibility. In the case of closed-source models,
                            it may be that access to the model is limited in
                            some way (e.g., to registered users), but it should be
                            possible for other researchers to have some path to reproducing
                            or verifying the results.
                    \end{itemize}
            \end{itemize}

        \item {\bf Open access to data and code}

        \item[] Question: Does the paper provide open access to the data and
            code, with sufficient instructions to faithfully reproduce the main experimental
            results, as described in supplemental material?

        \item[] Answer: \answerYes{} 

        \item[] Justification: We have open-sourced codes and model checkpoints.

        \item[] Guidelines:
            \begin{itemize}
                \item The answer NA means that paper does not include experiments
                    requiring code.

                \item Please see the NeurIPS code and data submission guidelines
                    (\url{https://nips.cc/public/guides/CodeSubmissionPolicy}) for
                    more details.

                \item While we encourage the release of code and data, we
                    understand that this might not be possible, so “No” is an acceptable
                    answer. Papers cannot be rejected simply for not including
                    code, unless this is central to the contribution (e.g., for
                    a new open-source benchmark).

                \item The instructions should contain the exact command and environment
                    needed to run to reproduce the results. See the NeurIPS code
                    and data submission guidelines (\url{https://nips.cc/public/guides/CodeSubmissionPolicy})
                    for more details.

                \item The authors should provide instructions on data access and
                    preparation, including how to access the raw data, preprocessed
                    data, intermediate data, and generated data, etc.

                \item The authors should provide scripts to reproduce all experimental
                    results for the new proposed method and baselines. If only a
                    subset of experiments are reproducible, they should state which
                    ones are omitted from the script and why.

                \item At submission time, to preserve anonymity, the authors should
                    release anonymized versions (if applicable).

                \item Providing as much information as possible in supplemental material
                    (appended to the paper) is recommended, but including URLs
                    to data and code is permitted.
            \end{itemize}

        \item {\bf Experimental setting/details}

        \item[] Question: Does the paper specify all the training and test
            details (e.g., data splits, hyperparameters, how they were chosen,
            type of optimizer, etc.) necessary to understand the results?

        \item[] Answer: \answerYes{} 

        \item[] Justification: Section~\ref{sec:setup-cifar10} and
            \ref{sec:setup-tinystories} give detailed settings of our
            experiments.

        \item[] Guidelines:
            \begin{itemize}
                \item The answer NA means that the paper does not include experiments.

                \item The experimental setting should be presented in the core of
                    the paper to a level of detail that is necessary to appreciate
                    the results and make sense of them.

                \item The full details can be provided either with the code, in
                    appendix, or as supplemental material.
            \end{itemize}

        \item {\bf Experiment statistical significance}

        \item[] Question: Does the paper report error bars suitably and
            correctly defined or other appropriate information about the
            statistical significance of the experiments?

        \item[] Answer: \answerYes{} 

        \item[] Justification: We report various of appropriate metrics in
            experiments.

        \item[] Guidelines:
            \begin{itemize}
                \item The answer NA means that the paper does not include experiments.

                \item The authors should answer "Yes" if the results are accompanied
                    by error bars, confidence intervals, or statistical significance
                    tests, at least for the experiments that support the main
                    claims of the paper.

                \item The factors of variability that the error bars are capturing
                    should be clearly stated (for example, train/test split,
                    initialization, random drawing of some parameter, or overall
                    run with given experimental conditions).

                \item The method for calculating the error bars should be explained
                    (closed form formula, call to a library function, bootstrap,
                    etc.)

                \item The assumptions made should be given (e.g., Normally
                    distributed errors).

                \item It should be clear whether the error bar is the standard deviation
                    or the standard error of the mean.

                \item It is OK to report 1-sigma error bars, but one should
                    state it. The authors should preferably report a 2-sigma error
                    bar than state that they have a 96\% CI, if the hypothesis of
                    Normality of errors is not verified.

                \item For asymmetric distributions, the authors should be
                    careful not to show in tables or figures symmetric error
                    bars that would yield results that are out of range (e.g.
                    negative error rates).

                \item If error bars are reported in tables or plots, The authors
                    should explain in the text how they were calculated and
                    reference the corresponding figures or tables in the text.
            \end{itemize}

        \item {\bf Experiments compute resources}

        \item[] Question: For each experiment, does the paper provide sufficient
            information on the computer resources (type of compute workers, memory,
            time of execution) needed to reproduce the experiments?

        \item[] Answer: \answerYes{} 

        \item[] Justification: FLOPs and memory usage are reported.

        \item[] Guidelines:
            \begin{itemize}
                \item The answer NA means that the paper does not include experiments.

                \item The paper should indicate the type of compute workers CPU or
                    GPU, internal cluster, or cloud provider, including relevant
                    memory and storage.

                \item The paper should provide the amount of compute required for
                    each of the individual experimental runs as well as estimate
                    the total compute.

                \item The paper should disclose whether the full research project
                    required more compute than the experiments reported in the paper
                    (e.g., preliminary or failed experiments that didn't make it
                    into the paper).
            \end{itemize}

        \item {\bf Code of ethics}

        \item[] Question: Does the research conducted in the paper conform, in every
            respect, with the NeurIPS Code of Ethics
            \url{https://neurips.cc/public/EthicsGuidelines}?

        \item[] Answer: \answerYes{} 

        \item[] Justification: We follow the rules of NeurIPS Code of Ethics.

        \item[] Guidelines:
            \begin{itemize}
                \item The answer NA means that the authors have not reviewed the
                    NeurIPS Code of Ethics.

                \item If the authors answer No, they should explain the special
                    circumstances that require a deviation from the Code of
                    Ethics.

                \item The authors should make sure to preserve anonymity (e.g.,
                    if there is a special consideration due to laws or
                    regulations in their jurisdiction).
            \end{itemize}

        \item {\bf Broader impacts}

        \item[] Question: Does the paper discuss both potential positive
            societal impacts and negative societal impacts of the work performed?

        \item[] Answer: \answerYes{} 

        \item[] Justification: We provide discussions in Appendix~\ref{supp:discussion}
            and \ref{supp:future-work}.

        \item[] Guidelines:
            \begin{itemize}
                \item The answer NA means that there is no societal impact of the
                    work performed.

                \item If the authors answer NA or No, they should explain why
                    their work has no societal impact or why the paper does not
                    address societal impact.

                \item Examples of negative societal impacts include potential malicious
                    or unintended uses (e.g., disinformation, generating fake profiles,
                    surveillance), fairness considerations (e.g., deployment of technologies
                    that could make decisions that unfairly impact specific groups),
                    privacy considerations, and security considerations.

                \item The conference expects that many papers will be foundational
                    research and not tied to particular applications, let alone
                    deployments. However, if there is a direct path to any
                    negative applications, the authors should point it out. For
                    example, it is legitimate to point out that an improvement in
                    the quality of generative models could be used to generate deepfakes
                    for disinformation. On the other hand, it is not needed to point
                    out that a generic algorithm for optimizing neural networks could
                    enable people to train models that generate Deepfakes faster.

                \item The authors should consider possible harms that could arise
                    when the technology is being used as intended and functioning
                    correctly, harms that could arise when the technology is
                    being used as intended but gives incorrect results, and harms
                    following from (intentional or unintentional) misuse of the technology.

                \item If there are negative societal impacts, the authors could
                    also discuss possible mitigation strategies (e.g., gated release
                    of models, providing defenses in addition to attacks, mechanisms
                    for monitoring misuse, mechanisms to monitor how a system
                    learns from feedback over time, improving the efficiency and
                    accessibility of ML).
            \end{itemize}

        \item {\bf Safeguards}

        \item[] Question: Does the paper describe safeguards that have been put
            in place for responsible release of data or models that have a high
            risk for misuse (e.g., pretrained language models, image generators,
            or scraped datasets)?

        \item[] Answer: \answerYes{} 

        \item[] Justification: No risks.

        \item[] Guidelines:
            \begin{itemize}
                \item The answer NA means that the paper poses no such risks.

                \item Released models that have a high risk for misuse or dual-use
                    should be released with necessary safeguards to allow for controlled
                    use of the model, for example by requiring that users adhere
                    to usage guidelines or restrictions to access the model or
                    implementing safety filters.

                \item Datasets that have been scraped from the Internet could pose
                    safety risks. The authors should describe how they avoided
                    releasing unsafe images.

                \item We recognize that providing effective safeguards is challenging,
                    and many papers do not require this, but we encourage authors
                    to take this into account and make a best faith effort.
            \end{itemize}

        \item {\bf Licenses for existing assets}

        \item[] Question: Are the creators or original owners of assets (e.g., code,
            data, models), used in the paper, properly credited and are the
            license and terms of use explicitly mentioned and properly respected?

        \item[] Answer: \answerYes{} 

        \item[] Justification: All the codes have been developed under the
            correct licenses.

        \item[] Guidelines:
            \begin{itemize}
                \item The answer NA means that the paper does not use existing assets.

                \item The authors should cite the original paper that produced the
                    code package or dataset.

                \item The authors should state which version of the asset is used
                    and, if possible, include a URL.

                \item The name of the license (e.g., CC-BY 4.0) should be included
                    for each asset.

                \item For scraped data from a particular source (e.g., website),
                    the copyright and terms of service of that source should be
                    provided.

                \item If assets are released, the license, copyright information,
                    and terms of use in the package should be provided. For popular
                    datasets, \url{paperswithcode.com/datasets} has curated
                    licenses for some datasets. Their licensing guide can help determine
                    the license of a dataset.

                \item For existing datasets that are re-packaged, both the
                    original license and the license of the derived asset (if it
                    has changed) should be provided.

                \item If this information is not available online, the authors
                    are encouraged to reach out to the asset's creators.
            \end{itemize}

        \item {\bf New assets}

        \item[] Question: Are new assets introduced in the paper well documented
            and is the documentation provided alongside the assets?

        \item[] Answer: \answerYes{} 

        \item[] Justification: All assets are released with comprehensive
            documents.

        \item[] Guidelines:
            \begin{itemize}
                \item The answer NA means that the paper does not release new assets.

                \item Researchers should communicate the details of the dataset/code/model
                    as part of their submissions via structured templates. This
                    includes details about training, license, limitations, etc.

                \item The paper should discuss whether and how consent was obtained
                    from people whose asset is used.

                \item At submission time, remember to anonymize your assets (if applicable).
                    You can either create an anonymized URL or include an anonymized
                    zip file.
            \end{itemize}

        \item {\bf Crowdsourcing and research with human subjects}

        \item[] Question: For crowdsourcing experiments and research with human
            subjects, does the paper include the full text of instructions given
            to participants and screenshots, if applicable, as well as details about
            compensation (if any)?

        \item[] Answer: \answerNA{} 

        \item[] Justification: There are no human subjects in the paper.

        \item[] Guidelines:
            \begin{itemize}
                \item The answer NA means that the paper does not involve crowdsourcing
                    nor research with human subjects.

                \item Including this information in the supplemental material is
                    fine, but if the main contribution of the paper involves
                    human subjects, then as much detail as possible should be included
                    in the main paper.

                \item According to the NeurIPS Code of Ethics, workers involved
                    in data collection, curation, or other labor should be paid
                    at least the minimum wage in the country of the data
                    collector.
            \end{itemize}

        \item {\bf Institutional review board (IRB) approvals or equivalent for research with human subjects}

        \item[] Question: Does the paper describe potential risks incurred by
            study participants, whether such risks were disclosed to the subjects,
            and whether Institutional Review Board (IRB) approvals (or an equivalent
            approval/review based on the requirements of your country or institution)
            were obtained?

        \item[] Answer: \answerNA{} 

        \item[] Justification: There are no human subjects in the paper.

        \item[] Guidelines:
            \begin{itemize}
                \item The answer NA means that the paper does not involve crowdsourcing
                    nor research with human subjects.

                \item Depending on the country in which research is conducted,
                    IRB approval (or equivalent) may be required for any human
                    subjects research. If you obtained IRB approval, you should
                    clearly state this in the paper.

                \item We recognize that the procedures for this may vary significantly
                    between institutions and locations, and we expect authors to
                    adhere to the NeurIPS Code of Ethics and the guidelines for
                    their institution.

                \item For initial submissions, do not include any information
                    that would break anonymity (if applicable), such as the institution
                    conducting the review.
            \end{itemize}

        \item {\bf Declaration of LLM usage}

        \item[] Question: Does the paper describe the usage of LLMs if it is an
            important, original, or non-standard component of the core methods
            in this research? Note that if the LLM is used only for writing,
            editing, or formatting purposes and does not impact the core methodology,
            scientific rigorousness, or originality of the research, declaration
            is not required.

        \item[] Answer: \answerNA{} 

        \item[] Justification: We did not use LLMs for developing our network.

        \item[] Guidelines:
            \begin{itemize}
                \item The answer NA means that the core method development in this
                    research does not involve LLMs as any important, original, or
                    non-standard components.

                \item Please refer to our LLM policy (\url{https://neurips.cc/Conferences/2025/LLM})
                    for what should or should not be described.
            \end{itemize}
    \end{itemize}
\end{document}